\documentclass[conference]{IEEEtran}
\IEEEoverridecommandlockouts
\usepackage[utf8]{inputenc} 
\usepackage{url}            
\usepackage{booktabs, tabularx}       
\usepackage{amsfonts}       
\usepackage{nicefrac}       
\usepackage{microtype}      
\usepackage{xcolor}         
\usepackage{soul} 
\usepackage{subcaption}
\usepackage{amsmath}
\usepackage{newfloat}
\usepackage{listings}
\usepackage{graphicx}
\usepackage{epstopdf}

\usepackage{pifont}

\usepackage{epsfig}
\usepackage{amsmath}
\usepackage{amsthm}
\usepackage{amsfonts}
\usepackage{bbding}
\usepackage{array}
\usepackage{paralist}
\usepackage{dsfont}
\usepackage{xargs}                      
\usepackage{caption}
\usepackage{ifthen}  

\newcolumntype{C}[1]{>{\centering\let\newline\\\arraybackslash\hspace{0pt}}m{#1}}

\usepackage{xcolor}

\newcommand{\dna}[1]{{\color{black}#1}}

\newcommand{\zero}[1]{{\boldsymbol{0}}}

\usepackage{multirow}


\usepackage{algorithm}
\usepackage{algpseudocode}

\errorcontextlines\maxdimen

\makeatletter
    \newcommand*{\algrule}[1][\algorithmicindent]{\makebox[#1][l]{\hspace*{.5em}\thealgruleextra\vrule height \thealgruleheight depth \thealgruledepth}}%
\newcommand*{\thealgruleextra}{}
\newcommand*{\thealgruleheight}{.75\baselineskip}
\newcommand*{\thealgruledepth}{.25\baselineskip}

\newcount\ALG@printindent@tempcnta
\def\ALG@printindent{%
    \ifnum \theALG@nested>0
        \ifx\ALG@text\ALG@x@notext
        \else
            \unskip
            \addvspace{-1pt}
            \ALG@printindent@tempcnta=1
            \loop
                \algrule[\csname ALG@ind@\the\ALG@printindent@tempcnta\endcsname]%
                \advance \ALG@printindent@tempcnta 1
            \ifnum \ALG@printindent@tempcnta<\numexpr\theALG@nested+1\relax
            \repeat
        \fi
    \fi
    }%
\usepackage{etoolbox}
\patchcmd{\ALG@doentity}{\noindent\hskip\ALG@tlm}{\ALG@printindent}{}{\errmessage{failed to patch}}
\makeatother

\newbox\statebox
\newcommand{\myState}[1]{%
    \setbox\statebox=\vbox{#1}%
    \edef\thealgruleheight{\dimexpr \the\ht\statebox+1pt\relax}%
    \edef\thealgruledepth{\dimexpr \the\dp\statebox+1pt\relax}%
    \ifdim\thealgruleheight<.75\baselineskip
        \def\thealgruleheight{\dimexpr .75\baselineskip+1pt\relax}%
    \fi
    \ifdim\thealgruledepth<.25\baselineskip
        \def\thealgruledepth{\dimexpr .25\baselineskip+1pt\relax}%
    \fi
    \State #1%
    \def\thealgruleheight{\dimexpr .75\baselineskip+1pt\relax}%
    \def\thealgruledepth{\dimexpr .25\baselineskip+1pt\relax}%
}

\AtBeginDocument{%
  \providecommand\BibTeX{{%
    \normalfont B\kern-0.5em{\scshape i\kern-0.25em b}\kern-0.8em\TeX}}}

\begin{document}

\title{Improving Time Series Encoding with Noise-Aware Self-Supervised Learning and an Efficient Encoder}
\author{\IEEEauthorblockN{1\textsuperscript{st} Duy A. Nguyen}
\IEEEauthorblockA{\textit{University of Illinois Urbana - Champaign} \\
\textit{VinUni-Illinois Smart Health Center}\\
Illinois, USA \\
duyan2@illinois.edu}
\and
\IEEEauthorblockN{2\textsuperscript{nd} Trang H. Tran}
\IEEEauthorblockA{\textit{Cornell University} \\
New York, USA \\
htt27@cornell.edu}
\and
\IEEEauthorblockN{3\textsuperscript{rd} Huy Hieu Pham}
\IEEEauthorblockA{\textit{VinUni-Illinois Smart Health Center}\\
\textit{College of Engineering \& Computer Science, VinUniversity} \\
Hanoi, Vietnam \\
hieu.ph@vinuni.edu.vn}
\and
\IEEEauthorblockN{4\textsuperscript{th} Phi Le Nguyen}
\IEEEauthorblockA{\textit{Hanoi University of Science and Technology} \\
Hanoi, Vietnam \\
lenp@soict.hust.edu.vn}
\and
\IEEEauthorblockN{5\textsuperscript{th} Lam M. Nguyen*}
\IEEEauthorblockA{\textit{IBM Research, Thomas J. Watson Research Center} \\
New York, USA \\
LamNguyen.MLTD@ibm.com \\
\textit{*Corresponding Author}
}
}

\maketitle
\thispagestyle{plain}
\pagestyle{plain}

\begin{abstract}
In this work, we investigate the time series representation learning problem using self-supervised techniques. Contrastive learning is well-known in this area as it is a powerful method for extracting information from the series and generating task-appropriate representations. Despite its proficiency in capturing time series characteristics, these techniques often overlook a critical factor - the inherent noise in this type of data, a consideration usually emphasized in general time series analysis. Moreover, there is a notable absence of attention to developing efficient yet lightweight encoder architectures, with an undue focus on delivering contrastive losses.
Our work address these gaps by proposing an innovative training strategy that promotes consistent representation learning, accounting for the presence of noise-prone signals in natural time series. Furthermore, we propose an encoder architecture that incorporates dilated convolution within the Inception block, resulting in a scalable and robust network with a wide receptive field. Experimental findings underscore the effectiveness of our method, consistently outperforming state-of-the-art approaches across various tasks, including forecasting, classification, and abnormality detection. Notably, our method attains the top rank in over two-thirds of the classification UCR datasets, utilizing only $40\%$ of the parameters compared to the second-best approach. 
\end{abstract}

\section{Introduction}\label{sec_intro}
\textbf{Motivation.}
Time series data, prevalent in fields like finance, medicine, and engineering, demand critical analysis for practical applications \cite{book1}. However, labeling such data is often challenging and expensive due to their complex and uninterpretable patterns, especially in sensitive domains such as healthcare and finance \cite{ts_challenge1, ts_challenge2}. Unsupervised learning offers a solution by enabling the acquisition of informative representations for various downstream tasks \dna{without the need for labels}.
Building upon the successes of similar techniques in computer vision and natural language processing \cite{cv1, nlp3}, there have been studies focusing on learning time series representations in an unsupervised manner \cite{t_loss, tnc, ts_tcc, btsf, ts2vec}. Despite significant progress, two notable gaps persist in the current literature: (1) the failure to explicitly address the inherent noisy characteristics of time series signals, and (2) the absence of effective and efficient encoder architectures tailored for processing such one-dimensional data. In this study, our goal is to introduce a versatile framework capable of simultaneously addressing both deficiencies within unsupervised settings.
\dna{Our design follows two crucial principles}: efficiency (ensuring accurate downstream task performance by capturing essential time series characteristics) and scalability (being lightweight to handle practical, lengthy, high-dimensional, and high-frequency time series data).
\begin{figure*}[!ht]
    \centering
    \includegraphics[width=0.8\textwidth]{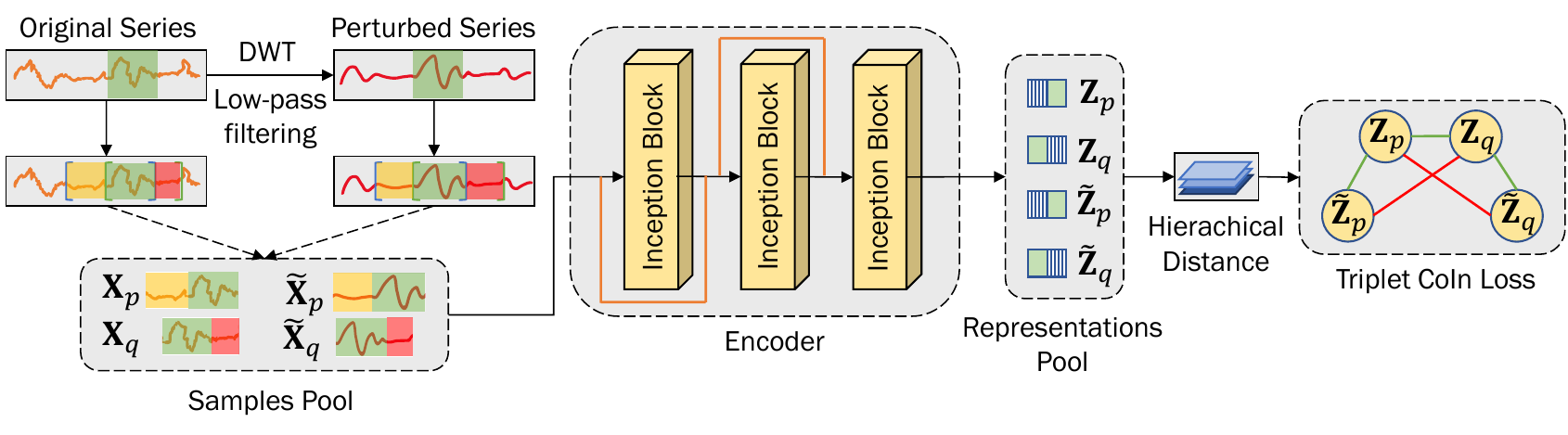}
    \caption{\textbf{Overview of the proposed CoInception framework}. We use samples from \textit{Representation Pool} in different downstream tasks.}
    \label{fig:overview}
\end{figure*}

\textbf{Related literature and our approach.}
Prior research on representation learning in time series data has predominantly focused on employing the self-supervised contrastive learning technique \cite{t_loss, tnc, ts_tcc, btsf, ts2vec}, which consists of two main components: \emph{training strategy} and \emph{encoder architecture}.

Existing training strategies revolve around time series' \emph{invariance characteristics}, encompassing temporal invariance \cite{cloc, tnc, t_loss}, transformation and augmentation invariance \cite{sim_crl, btsf, tf_c}, and contextual invariance \cite{ts_tcc, ts2vec}. For instance, TNC \cite{tnc} leverages temporal invariance for positive pair sampling but faces limitations in real-time applicability due to quadratic complexity. BTSF \cite{btsf} combines dropout and spectral representations, yet its efficiency relies on dropout rate and time instance length. Some studies \cite{ts_tcc, ts2vec} maintain contextual invariance, with \cite{ts2vec} focusing on consistent representations across different contexts (i.e., time segments). However, this may risk losing surrounding context information due to temporal masking.
A common deficiency in existing unsupervised methods is their treatment of noise during the learning of time series representations, a topic extensively explored in traditional time series analysis literature \cite{noise_1, noise_2}. Most of these methods either overlook the noisy nature of time series data or implicitly rely on Neural Networks' ability to handle such undesired signals, rather than explicitly addressing them during representation learning. This shortcoming has been shown to have a detrimental effect on tasks' accuracy \cite{song2022robust, wen2019robuststl}.
In recognition of this issue, we propose a training strategy guided by the principle that the presence of noise in the time series should not impede the functionality of our framework. Ideally, it should generate consistent representations whether provided with noise-free or raw series, highlighting \emph{noise-resiliency characteristics}.
To achieve this, we employ a spectrum-based low-pass filter to generate correlated yet distinct views of each input time series. The corresponding representations are then guided by our proposed system of loss functions. These loss functions effectively align embeddings of the raw-augmented couplets to attain desired noise invariance, while simultaneously preserving important information through a Triplet-based regularization term.
The advantages of this combination are twofold: (1) the filter preserves key characteristics such as trend and seasonality, ensuring deterministic and interpretable representations, while eliminating noise-prone high-frequency components; (2) the loss system stably directs the network in improving noise resilience and retaining information, leading to a significant enhancement in downstream task performance.

In addition to effective training strategies, the advancement of robust encoder architectures for generating versatile time series representations is frequently overshadowed by the former, which tends to attract more attention from researchers.
Common methods include linear models \cite{dlinear}, auto-encoders \cite{ts_autoend}, sequence-to-sequence models \cite{ts_seq2seq, ts_seq2seq2}, and Convolution-based designs like Causal Convolution \cite{ts_causal1, t_loss} and Dilated Convolution \cite{t_loss, ts2vec}. Yet, these approaches may struggle with long-term dependencies, particularly for extensive time series data. Alternatively, Transformer-based models and their variations \cite{tedformer, inparformer, patchtst} are adopted to address long-term dependencies, but can be computationally demanding and vulnerable to collapse on specific tasks or data \cite{transformer_collapse,transformer_collapse1}.
In response, we propose an efficient and scalable encoder framework, combining the strengths of Dilated Convolution and Inception idea. While Dilated Convolution achieves a broad receptive field without excessive depth, the Inception concept, which utilize multi-scale filters, effectively automate the process of choosing dilation factors, captures sequential correlations across scales. This dual approach balances representation effectiveness and model scalability. In addition, we enhance the vanilla Inception idea by introducing simple yet effective convolution-based aggregator and extra skip connections within the Inception block, boosting its ability to capture long-term dependencies in input time series.

\textbf{Our contributions.}
In this study, we introduce \emph{CoInception}, a noise-resilient, robust, and flexible representation learning framework for time series. Our main contributions are as follows.
\begin{itemize}
\item We directly address the adverse effects of noise in learning time series representations under unsupervised settings. Specifically, we introduce an effective training strategy encompassing combination of noise-resilient sampling step and loss system that enables learning consistent representations even in the presence of noise in natural time series data.
\item We present a robust and scalable encoder that leverages the advantages of well-established Inception blocks and the Dilation concept in convolution layers. With this, we can maintain a lightweight, shallow, yet robust framework while ensuring a wide receptive field of the final output. 
\item We conducted thorough experiments to assess the effectiveness of CoInception and examine its behavior. Our empirical findings indicate that our approach surpasses the current state-of-the-art methods across three primary time series tasks: forecasting, classification, and anomaly detection. Furthermore, extensive analysis reveals the potential of CoInception when combined with various approaches in diverse scenarios to improve overall performance.
\end{itemize}

\section{Proposed Method}
In this section, we majorly describe the working of the CoInception framework. 
We first present mathematical definitions of different time series problems in Sec. \ref{subsec:problem}. Following, the technical details and training methodology of our method would be discussed in Sec. \ref{subsec:framework} and \ref{subsec:loss}.
\begin{figure*}[hbt]
    \centering
    \includegraphics[width=0.8\textwidth]{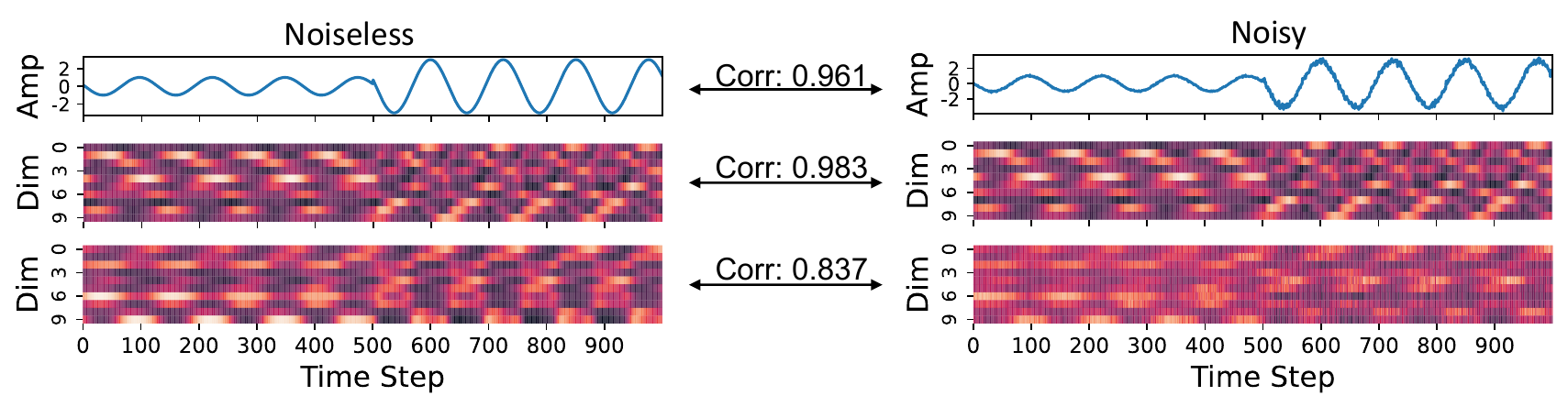}
    \caption{\textbf{Output representations for toy time series containing high-frequency noise.} CoInception's results can still capture the periodic characteristics regardless of the presence of noise.}
    \label{fig:trajectory_toy}
\end{figure*}
\subsection{Problem Formulation}
\label{subsec:problem}
The majority of natural time series can be represented as a continuous or discrete stream. Without loss of generality, we only consider the discrete series (continuous ones can be discretized through a quantization process). 

Let $\mathcal{X}=\left\{\mathbf{x}_1, \mathbf{x}_2, \dots, \mathbf{x}_n \right\}$ be such a dataset with $n$ sequences, where $\mathbf{x}_i \in \mathbb{R}^{M \times N}$ ($M$ is sequence length and $N$ is number of features), our goal is to obtain the corresponding latent representations $\mathcal{Z} = \left\{\mathbf{z}_1, \mathbf{z}_2, \dots, \mathbf{z}_n \right\}$, in which $\mathbf{z}_i \in \mathbb{R}^{M \times H}$ ($M$ is sequence length and $H$ is desired latent dimension). The time resolution of the learnt representations is kept intact as the original sequences, which has been shown to be more beneficial for adapting the representations to many downstream tasks \cite{ts2vec}. Our ultimate goal of learning the latent representations is to adapt them as the new input for popular time series tasks, defined by distinguished objectives. 
Let $\mathbf{z}_i = \left[z^1_i, \dots, z^M_i\right]$ be the learned representation for each segment, we can describe those objectives as follow. 
\begin{itemize}
    \item \textbf{Forecasting} requires the prediction of corresponding $T$-step ahead future observations $\mathbf{y}_i = \left[y^{M+1}_i, \dots, y^{M+T}_i\right]$; 
    
    \item \textbf{Classification} aims at identifying the correct label in the form $\mathbf{p}_i = \left[p_{1}, \dots, p_{C}\right]$, where $C$ is the number of classes;
    
    \item \textbf{Anomaly detection } determines whether the last time step $x^M_i$ (corresponding to $z^M_i$) is an abnormal point (streaming evaluation protocol - \cite{ren2019time}). 
\end{itemize}
From now on, without further mention, we would implicitly exclude the index number $i$ for readability.

\subsection{CoInception Framework}
\label{subsec:framework}
Adopting an unsupervised contrastive learning strategy, CoInception framework can be decomposed into three distinct components: (1) Sampling step, (2) Encoder architecture, and (3) Loss function. Figure \ref{fig:overview} illustrates the overall architecture.

\subsubsection{Sampling Strategy}
\label{sec:sampling_strategy}
Natural time series often contain noise, represented as a random process that oscillates independently alongside the main signal (e.g., white noise \cite{white_noise1}). To illustrate, a series $\mathbf{x}$ can be decomposed into distinct components $\mathbf{\hat{x}}$ and $\mathbf{n}$, representing the original signal and an independent noise component, respectively. While existing methods typically treat the signal and noise separately, even when only the noise factor varies (e.g., $\mathbf{n} \rightarrow \mathbf{\Tilde{n}}$), we contend that high-frequency noise-like elements, which are prominent in the high-frequency spectrum of the original series, contribute little to no meaningful information and can greatly degrade the accuracy of downstream tasks. This realization aligns with prior studies \cite{last, cost} emphasizing the importance of utilizing distinct components of raw series, such as seasonality or trends, which exhibit long-term persistence and are present within the low-frequency spectrum. Consequently, we underscore the importance of noise resilience in representations, enabling them to withstand such high-frequency signals.

Realizing such important characteristics, we first validate the sensitivity of existing frameworks with noise, by conducting a toy experiment with a synthesized series (upper left plot in Fig. \ref{fig:trajectory_toy}) and its disturbed version with two noise signals added (upper right plot of Fig. \ref{fig:trajectory_toy}). We adopt cosine similarity as the correlation measurement. Considering the high correlation ($0.961$) between noisy and noiseless series, together with their negligible visual differences (Fig. \ref{fig:trajectory_toy}), we expect the fundamental characteristics of learnt representations to remain intact. However, an existing state-of-the-art (SOTA) framework - TS2Vec \cite{ts2vec} fails to exhibit such a strong relation (correlation reduced to $0.837$, visually demonstrated by two bottom trajectories), highlighting its noise susceptibility. In contrast, CoInception's outcomes (two middle trajectories) show strong consistence (correlation of $0.983$), capturing the original sine wave's harmonic shift even in noisy scenario.

\begin{figure*}[th]
    \centering
    \begin{subfigure}{0.43\textwidth}
    \centering
    \includegraphics[width=\textwidth]{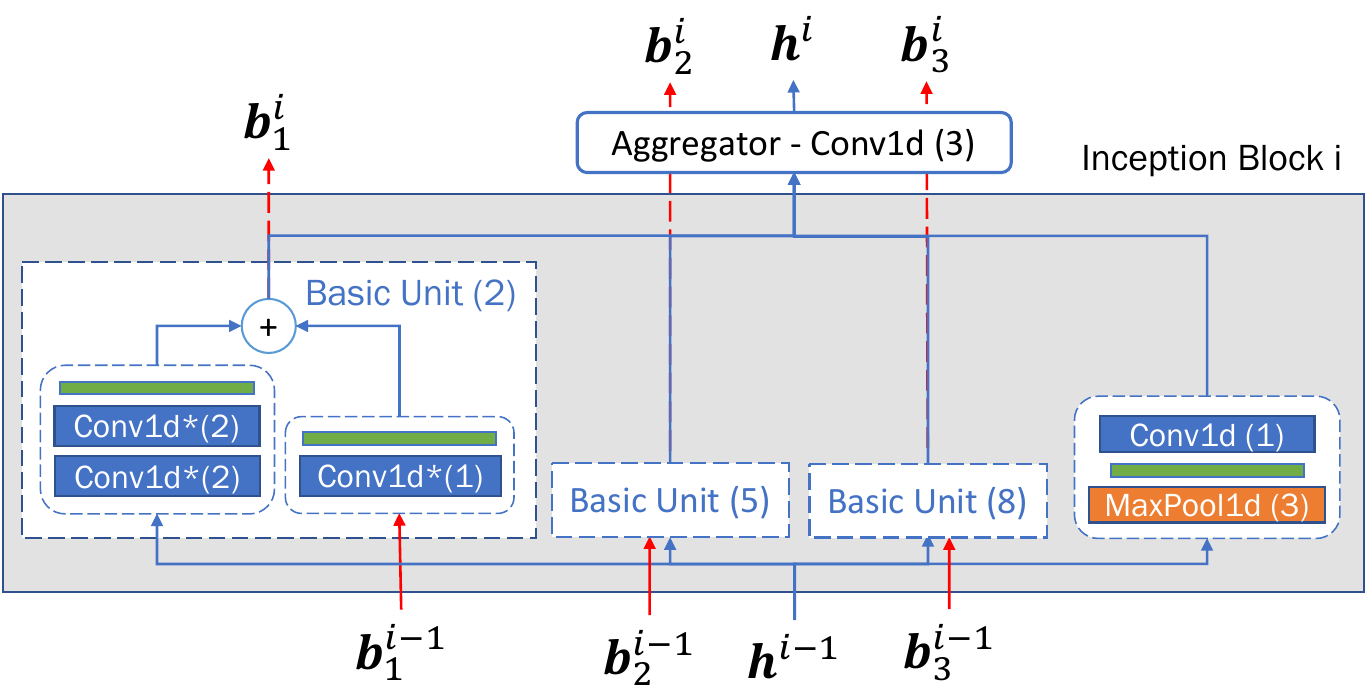}
    \caption{Inception Block}
    \label{fig:inception_block}
    \end{subfigure}%
    \hfill
    \begin{subfigure}{0.56\textwidth}
    \centering
        \raisebox{0.2\height}{\includegraphics[width=\textwidth]{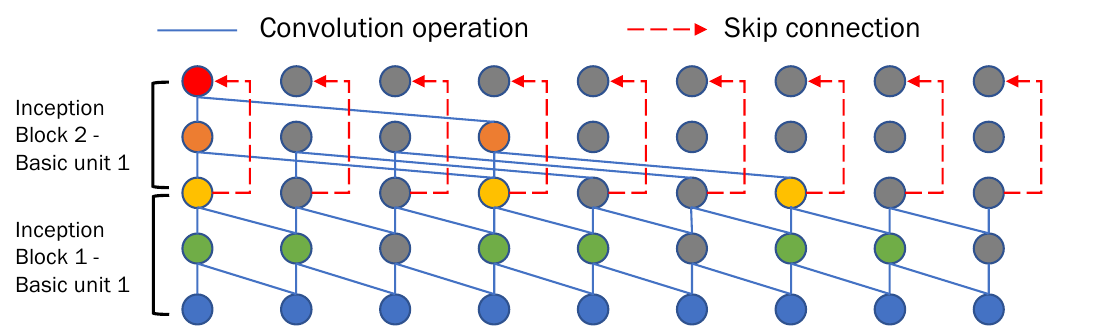}}
    \caption{Receptive field illustration}
    \label{fig:reception_field}
    \end{subfigure}
    \caption{Illustration of the Inception block and the accumulated receptive field upon stacking.}
\end{figure*}

Figure \ref{fig:overview} illustrates an overview of the proposed sampling strategy, which operates in conjunction with our proposed loss system to guarantee the noise resilience properties of the learned representations.
We leverage Discrete Wavelet Transform (DWT) as a parameter-free low-pass filter \cite{dwt} to generate a perturbed version $\mathbf{\Tilde{x}}$ of original series $\mathbf{x}$. The DWT filter convolves the input series with a set of shifted wavelet functions to generate coefficients representing their contributions at various intervals, before downsampling the result by the factor of $2$. This filter is applied $L$ times, corresponding to $L$ levels of decomposition, with the output of previous iteration be the input of the next one. This process essentially segregates the original series into $L+1$ distinct frequency bands, in which the low-frequency approximation coefficients reflect the overall trend of the data, whereas the high-frequency detail coefficients represent noise-like components. Here, $L = \lfloor \log_2(\tfrac{M}{K}) \rfloor$ is the maximum useful level of the decomposition, where $K$ is the length of chosen mother wavelet.
Mathematically, let $\mathbf{g}$ and $\mathbf{h}$ denote the low-pass and its quadrature-mirror high-pass filters, respectively. The working of DWT filter at the $j$-th level and position $n$ is as follows.
\begin{equation*}
\begin{aligned}
    \mathbf{x}^j[n] &= (\mathbf{x}^{j-1} * \mathbf{g}^{j})[n] = \sum_{k} \mathbf{x}^{j-1}[k] \mathbf{g}^j[2n-k], \\
    \mathbf{d}^j[n] &= (\mathbf{x}^{j-1} * \mathbf{h}^{j})[n] = \sum_{k} \mathbf{x}^{j-1}[k] \mathbf{h}^j[2n-k].
\end{aligned}
\end{equation*}
where $*$ represents convolution operator, $k$ denotes the shifted coefficient, $\mathbf{g}^{j}$ and $\mathbf{h}^{j}$ are the low-pass and high-pass filter coefficients, $\mathbf{x}^j$ and $\mathbf{d}^j$ represent the approximation and detail coefficients.
Following, to create a perturbed version of the input series, we intentionally retain only the significant values in the detail coefficients $\mathbf{d}^j$ ($j \in \left\{1, \dots, L \right\}$), while masking out unnecessary (potentially noise) values, which result in perturbed detail coefficients $\mathbf{\Tilde{d}}^j$ as follows.
\begin{equation}
\begin{aligned}
    \mathbf{\Tilde{d}}^j  = \left[ \frac{d_k}{|d_k|} \times \max(|d_k| - \gamma, 0) \Big| k \in \left\{ 1, \dots, \text{len}(\mathbf{d}^j) \right\} \right] .
\end{aligned}
\end{equation}
With this strategy, we define a cutting threshold $\gamma$ to be proportional to the maximum value of input series $\mathbf{x}$ by a hyper-parameter $\alpha < 1$, i.e., $\gamma = \alpha \times \max(\mathbf{x})$. 
Subsequently, the reconstruction process involves the approximation coefficient $\mathbf{\mathbf{x}^L}$ and set of perturbed detail coefficients $\left\{ \mathbf{\Tilde{d}}^1, \dots, \mathbf{\Tilde{d}}^L \right\}$ using the inverse Discrete Wavelet Transform (iDWT), producing the modified series $\mathbf{\Tilde{x}}$. 
The sampling phase concludes with the implementation of random cropping on both $\mathbf{x}$ and $\mathbf{\Tilde{x}}$, resulting in overlapping segments $\left<\mathbf{x}_p; \mathbf{x}_q\right>$ and $\left<\mathbf{\Tilde{x}}_p; \mathbf{\Tilde{x}}_q\right>$. These segments are subsequently utilized by the CoInception encoder (Section \ref{sec:encoder}).
\subsubsection{Inception-Based Dilated Convolution Encoder}
\label{sec:encoder}

In pursuit of an architecture that strikes a balance between robustness and efficiency, we deliver the CoInception encoder which integrates principles from Dilated Convolution and the Inception concept. Previous studies \cite{t_loss, ts2vec} highlight the robustness of stacked Dilated Convolutional Networks in various tasks, emphasizing their potential. The strength of this architecture lies in its ability to retain low scale of networks parameters, while maintaining robustness via a large accumulative receptive field. However, a key weakness arises in the selection of dilation factors, posing a trade-off between effectiveness and efficiency. Small factors reduce the parameter-efficient gain, while large factors risk focusing too much on broad contextual information, neglecting local details. To address this, our design utilizes the concept of Inception, which naturally automates the incorporation of different dilation factors into a single layer, constituting Inception block (Fig.\ref{fig:inception_block}). Specifically, within each block, there are several Basic units encompassing 1D convolutional layers of varying filter lengths and dilation factors. This configuration enables the encoder to consider input segments at diverse scales and resolutions.

In addition, apart from existing Inception-based models \cite{inception_ts}, \cite{timesnet}, we introduce two additional modifications to improve robustness and scalability, without sacrificing design simplicity: (1) an aggregator layer, and (2) extra skip connections (i.e., red arrows in Fig. \ref{fig:reception_field}).
Regarding the aggregator, beyond the aim of reducing the number of parameters as in \cite{inception_ts}, it is intentionally placed after the Basic units to better combine the features $\mathbf{b}$ produced by those layers, producing aggregated representation $\mathbf{h}$. Moreover, with the stacking nature of Inception blocks in our design, the aggregator can still inherit the low-channel-dimension output of the previous block, just like the conventional Bottleneck layer.
Furthermore, we introduce extra skip connections that interconnect the outputs of these units across different Inception blocks, denoted as modification (2). These skip connections have two-fold benefits of serving as shortcut links for stable gradient flow and gluing up the Basic units of different Inception blocks, making the entire encoder horizontally and vertically connected. 
In this way, our CoInception framework can be seen as a set of multiple Dilated Convolution experts, with much shallower depth and equivalent receptive fields compared with ordinary stacked Dilated Convolution networks \cite{t_loss, ts2vec}. 
Mathematically, let $k_u$ be the base kernel size (the numbers in bracket of Fig.\ref{fig:inception_block}) for a Basic unit $u$ within the $i^{th}$ Inception block ($1$-indexed), the dilation factor and the receptive field are calculated as $d_u^i = (2k - 1)^{i-1}; r_u^i = (2k - 1)^{i}$.
Illustrated in Figure \ref{fig:reception_field} is the accumulative receptive field associated with the Basic unit featuring a base kernel size of $2$, at the first and second Inception blocks.
\subsection{Hierarchical Triplet Loss}
\label{subsec:loss}
In conjuction with the sampling strategy outlined in Section \ref{sec:sampling_strategy}, a system of loss functions are deployed to attain robust and noise-resilient representation. We integrate the concept of hierarchical loss \cite{ts2vec} and triplet loss \cite{triplet_loss} to enhance \emph{noise resiliency}, incorporating a variation of \emph{contextual consistency} inspired by \cite{ts2vec}.
\dna{For simplicity in annotation, we use $\left<\mathbf{z}_p; \mathbf{z}_q \right>$ and $\left<\mathbf{\Tilde{z}}_p; \mathbf{\Tilde{z}}_q\right>$ to denote the representations of the actual overlapping segments between the sampled couplets $\left<\mathbf{x}_p; \mathbf{x}_q \right>$ and $\left<\mathbf{\Tilde{x}}_p; \mathbf{\Tilde{x}}_q\right>$ (green timestamps in Fig.\ref{fig:overview})}. 
With this, the \emph{noise-resilient} characteristic is ensured by minimizing the distances between representations of the original segments and their perturbed views - $\left<\mathbf{z}_p; \mathbf{\Tilde{z}}_p\right>$ and $\left<\mathbf{z}_q; \mathbf{\Tilde{z}}_q\right>$. In parallel, the embeddings $\mathbf{z}_p$ and $\mathbf{z}_q$ should also be close in latent space to preserve the \emph{contextual consistency}.
To model the distance within a couplet, we incorporate both instance-wise loss \cite{t_loss} ($\mathcal{L}_{ins}$) and temporal loss \cite{tnc} ($\mathcal{L}_{temp}$). The combination of these two forms the consistency loss $\mathcal{L}_{con}$. 
\begin{equation*}
    \begin{aligned}
        \mathcal{L}_{temp}(\mathbf{z}_p, \mathbf{z}_q) &= \tfrac{-1}{BT} \sum^{B,T}_{b,t} \log \tfrac{\exp(z^{b,t}_p \cdot z^{b,t}_q)}{\sum^T_{\Tilde{t}}\left(\exp(z^{b,t}_p \cdot z^{b,\Tilde{t}}_q) + \mathds{1}_{t \neq \Tilde{t}} \exp(z^{b,t}_p \cdot z^{b,\Tilde{t}}_p) \right)}, \\
        \mathcal{L}_{ins}(\mathbf{z}_p, \mathbf{z}_q) &= \tfrac{-1}{BT} \sum^{B,T}_{b,t} \log \tfrac{\exp(z^{b,t}_p \cdot z^{b,t}_q)}{\sum^B_{\Tilde{b}}\left(\exp(z^{b,t}_p \cdot z^{\Tilde{b}, t}_q) + \mathds{1}_{b \neq \Tilde{b}} \exp(z^{b,t}_p \cdot z^{\Tilde{b}, t}_p) \right)}, \\
        \mathcal{L}_{con}(\mathbf{z}_p, \mathbf{z}_q) &= \mathcal{L}_{temp}(\mathbf{z}_p, \mathbf{z}_q) + \mathcal{L}_{ins}(\mathbf{z}_p, \mathbf{z}_q).
    \end{aligned} 
\end{equation*}
In addition, to further enhance the reliability of learned representations, we propose to enforce an auxiliary criteria, based on the following observation. Apart from the previously mentioned pairs, extra couplets can be formed by comparing the representations of an original segment with a perturbed version in a different context, e.g., $\left<\mathbf{z}_p; \mathbf{\Tilde{z}}_q\right>$. \dna{$\mathbf{z}_p$ and $\mathbf{z}_q$ are from the same original samples, forming the common region of two segments $\mathbf{x}_p$ and $\mathbf{x}_q$. Conversely, $\mathbf{\Tilde{z}}_p$ or $\mathbf{\Tilde{z}}_q$ arises from the overlap of two perturbed segments $\mathbf{\Tilde{x}}_p$ and $\mathbf{\Tilde{x}}_q$. Therefore, it is reasonable to expect the proximity between $\mathbf{z}_p$ and the unaltered representation $\mathbf{z}_q$ is greater than that of $\mathbf{z}_p$ and the modified counterpart $\mathbf{\Tilde{z}}_q$.}
This condition could also help in mitigating the over-smoothing effect potentially caused by the DWT-low pass filter.
We incorporate this observation as a constraint in the final loss function (denoted as $\mathcal{L}_{triplet}$) in the format of a triplet loss as follows.
\begin{equation}
\begin{aligned}
    \mathcal{L}_{triplet}&(l_{pq}, l_{p\Tilde{q}}, l_{\Tilde{p}q}, \epsilon, \zeta) = \epsilon \times \tfrac{l_{pq} + l_{p\Tilde{p}} + l_{q\Tilde{q}}}{3} \\
    + &(1 - \epsilon) \times \max(0, 2 \times l_{pq} - l_{p\Tilde{q}} - l_{\Tilde{p}q} + 2\times \zeta),
\end{aligned}
\label{eq:loss}
\end{equation}
where $l_{pq}$ represents $\mathcal{L}_{con}(\mathbf{z}_p, \mathbf{z}_q)$, $l_{p\Tilde{q}}$ is $\mathcal{L}_{con}(\mathbf{z}_p, \mathbf{\Tilde{z}}_q)$ and similar notations for remaining terms. $\epsilon < 1$ is the balance factor for two loss terms, while $\zeta$ denotes the triplet margin. To ensure the CoInception framework can handle inputs of multiple granularity levels, we adopt a hierarchical strategy similar to \cite{ts2vec} with our $\mathcal{L}_{triplet}$ loss (Algorithm \ref{alg:hier_loss}).
\begin{algorithm}[h]
  \caption{Hierarchical Triplet Loss Calculation} \label{alg:hier_loss}
  \textbf{Input}: \\
    $\triangleright \hspace{0.2cm} \mathbf{z}_i, \mathbf{z}_j$ - embeddings of $i^{th}$ and $j^{th}$ segments; \\
    $\triangleright \hspace{0.2cm} \mathbf{\Tilde{z}}_i, \mathbf{\Tilde{z}}_j$ - embeddings of $i^{th}$ and $j^{th}$ \textbf{perturbed} segments; \\
    $\triangleright \hspace{0.2cm} \epsilon$ - Balance factor between instance loss and temporal loss; \\
    $\triangleright \hspace{0.2cm} \zeta$ - Triplet loss margin. \\
    \textbf{Output}: \\
    $\triangleright \hspace{0.2cm} l_{3hier}$ - Hierarchical triplet loss value
    \vspace{0.2em}
  \begin{algorithmic}[1]
    \Function{HierTripletLoss()}{}
      \myState{\textit{Initialize} $l_{3hier} \leftarrow 0$; $r \leftarrow 0$}
      \Comment{Running variable}
      \While{$\texttt{time\_dimension}(\mathbf{z}_i) > 1$} \\
      \Comment{Loop with reduced time resolution} 
        \myState{$l_{ij}, l_{i\Tilde{i}}, l_{j\Tilde{j}} \leftarrow \mathcal{L}_{con}(\mathbf{z}_i, \mathbf{z}_j), \mathcal{L}_{con}(\mathbf{z}_i, \mathbf{z}_{\Tilde{i}}), \mathcal{L}_{con}(\mathbf{z}_j, \mathbf{z}_{\Tilde{j}})$;} \\
        \Comment{Losses for main couplets} \\
        \myState{$l_{i\Tilde{j}}, l_{\Tilde{i}j} \leftarrow \mathcal{L}_{con}(\mathbf{z}_i, \mathbf{z}_{\Tilde{j}}), \mathcal{L}_{con}(\mathbf{z}_{\Tilde{i}}, \mathbf{z}_j)$;} \\
        \Comment{Losses for supporting couplets} \\
        \myState{$l_{3hier} \leftarrow l_{3hier} + \mathcal{L}_{triplet}(l_{ij}, l_{i\Tilde{j}}, l_{\Tilde{i}j}, \epsilon, \zeta) $;} \\ 
        
        \myState{$\mathbf{z}_i, \mathbf{z}_j \leftarrow \texttt{mp\_1d}(\mathbf{z}_i), \texttt{mp\_1d}(\mathbf{z}_j)$;} \\
        \myState{$\mathbf{\Tilde{z}}_i, \mathbf{\Tilde{z}}_j  \leftarrow  \texttt{mp\_1d}(\mathbf{\Tilde{z}}_i), \texttt{mp\_1d}(\mathbf{\Tilde{z}}_j)$;} \\
        \Comment{Reducing the time resolution} \\
        \myState{$d \leftarrow d + 1$}
  \EndWhile
    \myState{$l_{3hier} \leftarrow l_{3hier} / d$}
    \myState{\textbf{return} $l_{3hier}$}
    \EndFunction
  \end{algorithmic}
\end{algorithm}

\section{Experiments}\label{sec:experiment}
In this section, we empirically validate the effectiveness of the CoInception framework and compare the results with the recent state of the arts. We consider three major tasks, including forecasting, classification, and anomaly detection, as in Section \ref{subsec:problem}. In all of our experiments, we highlight best results in bold and \textcolor{red}{\textbf{red}}, and second best results are in \textcolor{blue}{blue}. 

\textbf{Baselines.} We majorly select methods following unsupervised training strategies and target multiple tasks: (1) \textbf{TS2Vec} \cite{ts2vec} learns to preserve contextual invariance across multiple time resolutions using a sampling strategy and hierarchical loss; (2) \textbf{TS-TCC} \cite{ts_tcc} combines cross-view prediction and contrastive learning tasks by creating two views of the raw time series data using weak and strong augmentations; (3) \textbf{TNC} \cite{tnc} tailors for time series data that forms positive and negative pairs from nearby and distant segments, respectively, leveraging the stationary properties of time series. 
Additionally, we include a recent work - (4) \textbf{TimesNet} \cite{timesnet} for comparison. TimesNet utilizes the multi-periodicity in time series to capture their temporal variations and has shown effectiveness in various practical tasks. To ensure a fair comparison, we adopt TimesNet to an unsupervised setting using the commonly used reconstruction task \cite{reconstruct_1, reconstruct_2}, and denote this modified version as TimesNet*.

\textbf{Hardware.} All implementations and experiments are performed on a single machine with the following hardware configuration: an $64-$core Intel Xeon CPU with a GeForce RTX 3090 GPU to accelerate training. 

\subsection{Time-Series Forecasting}  
\renewcommand{\arraystretch}{1.0}
\begin{table*}[!ht]
\setlength\doublerulesep{0.8pt}
\centering
\caption{Multivariate time series forecasting results on MSE. 
}
\label{tab:mulvar_forecasting_mse}
\centering
\setlength\tabcolsep{3pt} 
\resizebox{\textwidth}{!}{%
\begin{tabular}{cccccccccccccccc}
\toprule[1pt]\midrule[0.3pt]
T   & TS2Vec                                & TS-TCC & TNC   & Informer                     & StemGNN                               & TimesNet* & CoInception                           & T   & TS2Vec                       & TS-TCC                       & TNC                          & Informer                              & StemGNN & TimesNet*                     & CoInception                           \\ \cmidrule(lr){1-8} \cmidrule(lr){9-16}
\multicolumn{8}{l}{\textit{ETTh1:}}                                                                                                                                                    & \multicolumn{8}{l}{\textit{ETTm1:}}                                                                                                                                                                                       \\
24  & 0.599                                 & 0.653  & 0.632 & \textcolor{blue}{0.577} & 0.614                                 & 0.914    & \textcolor{red}{\textbf{0.461}} & 24  & 0.443                        & 0.473                        & 0.429                        & \textcolor{red}{\textbf{0.323}} & 0.620   & 1.005                        & \textcolor{blue}{0.384}          \\
48  & \textcolor{blue}{0.629}          & 0.720  & 0.705 & 0.685                        & 0.748                                 & 1.006    & \textcolor{red}{\textbf{0.512}} & 48  & 0.582                        & 0.671                        & 0.623                        & \textcolor{red}{\textbf{0.494}} & 0.744   & 1.008                        & \textcolor{blue}{0.552}          \\
168 & 0.755                                 & 1.129  & 1.097 & 0.931                        & \textcolor{red}{\textbf{0.663}} & 1.105    & \textcolor{blue}{0.683}          & 96  & \textcolor{blue}{0.622} & 0.803                        & 0.749                        & 0.678                                 & 0.709   & 1.104                        & \textcolor{red}{\textbf{0.561}} \\
336 & \textcolor{blue}{0.907}          & 1.492  & 1.454 & 1.128                        & 0.927                                 & 1.15    & \textcolor{red}{\textbf{0.829}} & 288 & \textcolor{blue}{0.709} & 1.958                        & 1.791                        & 1.056                                 & 0.843   & 1.109                        & \textcolor{red}{\textbf{0.623}} \\
720 & \textcolor{blue}{1.048}          & 1.603  & 1.604 & 1.215                        & -                                     & 1.348    & \textcolor{red}{\textbf{1.018}} & 672 & \textcolor{blue}{0.786} & 1.838                        & 1.822                        & 1.192                                 & -       & 1.115                        & \textcolor{red}{\textbf{0.717}} \\ 
\cmidrule(lr){1-8} \cmidrule(lr){9-16}
\multicolumn{8}{l}{\textit{ETTh2:}}                                                                                                                                                    & \multicolumn{8}{l}{\textit{Electricity:}}                                                                                                                                                                                 \\
24  & \textcolor{blue}{0.398}          & 0.883  & 0.830 & 0.720                        & 1.292                                 & 0.915    & \textcolor{red}{\textbf{0.335}} & 24  & 0.287                        & \textcolor{blue}{0.278} & 0.305                        & 0.312                                 & 0.439   & 0.414                        & \textcolor{red}{\textbf{0.234}} \\
48  & \textcolor{blue}{0.580}          & 1.701  & 1.689 & 1.457                        & 1.099                                 & 1.709    & \textcolor{red}{\textbf{0.550}} & 48  & \textcolor{blue}{0.307} & 0.313                        & 0.317                        & 0.392                                 & 0.413   & 0.516                        & \textcolor{red}{\textbf{0.265}} \\
168 & \textcolor{blue}{1.901}          & 3.956  & 3.792 & 3.489                        & 2.282                                 & 2.224    & \textcolor{red}{\textbf{1.812}} & 168 & \textcolor{blue}{0.332} & 0.338                        & 0.358                        & 0.515                                 & 0.506   & 0.585                        & \textcolor{red}{\textbf{0.282}} \\
336 & \textcolor{blue}{2.304}          & 3.992  & 3.516 & 2.723                        & 3.086                                 & 3.017    & \textcolor{red}{\textbf{2.151}} & 336 & \textcolor{blue}{0.349} & 0.357                        & \textcolor{blue}{0.349} & 0.759                                 & 0.647   & 0.601                        & \textcolor{red}{\textbf{0.301}} \\
720 & \textcolor{red}{\textbf{2.650}} & 4.732  & 4.501 & 3.467                        & -                                     & 3.121    & \textcolor{blue}{2.962}          & 720 & \textcolor{blue}{0.375}                        & 0.382                        & 0.447                        & 0.969                                 & -       & 0.663 & \textcolor{red}{\textbf{0.331}} \\ 
\midrule[0.3pt]\bottomrule[1pt]
\end{tabular}
}
\end{table*}

\begin{table*}[ht]
\begin{minipage}{0.45\linewidth}
\centering
        \includegraphics[width=1\columnwidth]{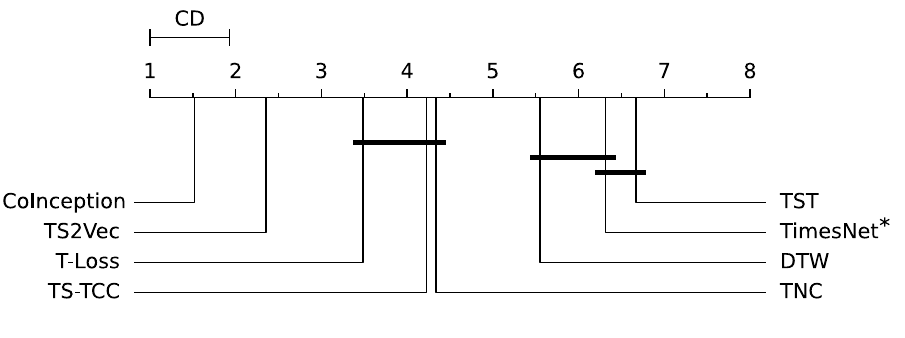}
        \captionof{figure}{Critical Difference Diagram comparing
different classifiers on 125 Datasets from UCR Repository with the
confidence level of $95\%$. } \label{fig:critical_diag}
\end{minipage}
\hfill
\begin{minipage}{0.54\linewidth}
\centering
\setlength\tabcolsep{4pt} 
\setlength\doublerulesep{0.8pt}
\captionof{table}{Time series classification results.} \label{tab:classification}
\resizebox{\columnwidth}{!}{
\begin{tabular}{lcccccc}
\toprule[1pt]\midrule[0.3pt]
\multirow{2}{*}{Dataset} & \multicolumn{3}{c}{UCR repository} & \multicolumn{3}{c}{UEA repository} \\ \cmidrule(lr){2-4} \cmidrule(lr){5-7} 
                          & Accuracy       & Rank          & Parameter     & Accuracy       & Rank          & Parameter     \\ \hline
DTW                      & 0.72           &  5.54          & -             & 0.65           & 3.96          & -             \\
TNC                      & 0.76           & 4.34          & -        & 0.68           & 4.60          &  -         \\
TST                      & 0.64           & 6.67          & 2.88M         & 0.64           & 5.66          & 2.88M         \\
TS-TCC                   & 0.76           & 4.22          & 1.44M         & 0.68           & 3.96          & 1.44M         \\
T-Loss                   & 0.81           & 3.48          & \textcolor{blue}{247K}          & 0.67           & 4.00          & \textcolor{blue}{247K}          \\
TimesNet* & 0.69 & 6.31 & 2.34M & 0.59 & 6.89 & 2.34M \\
TS2Vec                   & \textcolor{blue}{0.83}           & \textcolor{blue}{2.35}          & 641K          & \textcolor{blue}{0.71}           & \textcolor{blue}{3.03}          & 641K          \\
\textcolor{red}{\textbf{CoInception }  }          & \textcolor{red}{\textbf{0.84}} & \textcolor{red}{\textbf{1.51}} & \textcolor{red}{\textbf{206K}} & \textcolor{red}{\textbf{0.72}} & \textcolor{red}{\textbf{1.86}} & \textcolor{red}{\textbf{206K}} \\
\midrule[0.3pt]\bottomrule[1pt]
\end{tabular}}
\end{minipage}
\end{table*}

\textbf{Datasets \& Settings.} For this experiment, the same settings as \cite{informer} are adopted for both short-term and long-term forecasting. In addition to the representative works, CoInception is further compared with studies that delicately target the forecasting task, such as Informer \cite{informer}, StemGNN \cite{stemgnn} and N-BEATS \cite{nbeat}. Among these frameworks, Informer \cite{informer} is a supervised model which requires no extra regressor to process its produced representations. For other unsupervised benchmarks, a linear regression model is trained using the $L2$ norm penalty, with the learned representation $\mathbf{z}$ as input to directly predict future values. 
To ensure a fair comparison with works that only generate instance-level representations, only the $M^{th}$ timestep representation $z^M$ produced by the CoInception framework is used for the input segment. The evaluation of the forecast result is performed using two metrics, namely Mean Square Error (MSE) and Mean Absolute Error (MAE). For the datasets used, the Electricity Transformer Temperature (ETT) \cite{informer} datasets are adopted 
together with the UCI Electricity \cite{electricity} dataset.

\noindent \textbf{Results.}
Due to limited space, we only present the multivariate forecasting results on MSE in Table \ref{tab:mulvar_forecasting_mse}.
Apparently, the proposed CoInception framework achieves the best results in most scenarios over all 4 datasets in the multivariate setting. 
The numbers indicate that our method outperforms existing state-of-the-art methods in most cases. 
Furthermore, the Inception-based encoder design results in a CoInception model with only $40\%$ number of parameters compared with the second-best approach (see Table \ref{tab:classification}).
\begin{table*}[ht]
\setlength\doublerulesep{0.8pt}
\centering
\caption{Time series abnormaly detection results.}
\label{tab:abnormally_detection}
\centering
\scriptsize
\setlength\tabcolsep{3pt} 
\resizebox{\textwidth}{!}{%
\begin{tabular}{llcccccccccccc}
\toprule[1pt]\midrule[0.3pt]
\multirow{2}{*}{Dataset}                   & \multirow{2}{*}{Metrics} & \multicolumn{6}{c}{Normal Setting}                         & \multicolumn{6}{c}{Cold-start Setting}                            \\ \cmidrule(lr){3-8} \cmidrule(lr){9-14}
                                           &                          & SPOT  & DSPOT & TimesNet* & SR    & TS2Vec & \textcolor{red}{\textbf{CoInception}} & FFT   & Twitter-AD & TimesNet* & SR    & TS2Vec & \textcolor{red}{\textbf{CoInception}} \\ \midrule
\multicolumn{1}{c}{\multirow{3}{*}{Yahoo}} & F1        & 0.338 & 0.316 & 0.374 & 0.563 & \textcolor{blue}{0.745}  & \textcolor{red}{\textbf{0.769} }   & 0.291 & 0.245      & 0.273   & 0.529 & \textcolor{blue}{0.726}  & \textcolor{red}{\textbf{0.745} }   \\
\multicolumn{1}{c}{}                       & Precision & 0.269 & 0.241 & 0.519 & 0.451 & 0.729  & 0.790             & 0.202 & 0.166      & 0.431   & 0.404 & 0.692  & 0.733             \\
\multicolumn{1}{c}{}                       & Recall    & 0.454 & 0.458 & 0.292 & 0.747 & 0.762  & 0.748             & 0.517 & 0.462      & 0.199   & 0.765 & 0.763  & 0.754             \\ \midrule
\multirow{3}{*}{KPI}                       & F1        & 0.217 & 0.521 & 0.192 & 0.622 & \textcolor{blue}{0.677}  & \textcolor{red}{\textbf{0.681}  }  & 0.538 & 0.330      & 0.189   & 0.666 & \textcolor{blue}{0.676}  & \textcolor{red}{\textbf{0.682}}    \\
                                           & Precision & 0.786 & 0.623 & 0.493 & 0.647 & 0.929  & 0.933             & 0.478 & 0.411      & 0.443   & 0.637 & 0.907  & 0.893             \\
                                           & Recall    & 0.126 & 0.447 & 0.119 & 0.598 & 0.533  & 0.536             & 0.615 & 0.276      & 0.120   & 0.697 & 0.540  & 0.552             \\ 
                          \midrule[0.3pt]\bottomrule[1pt]
\end{tabular}
}
\end{table*}
\subsection{Time-Series Classification} 
\textbf{Datasets \& Settings.} For the classification task, we follow the settings in \cite{t_loss} and train an RBF SVM classifier on instance-level representations generated by our baselines. However, since CoInception produces timestamp-level representations for each data instance, we utilize the strategy from \cite{ts2vec} to ensure a fair comparison. Specifically, we apply a global MaxPooling operation over $\mathbf{z}$ to extract the instance-level vector representing the input segment. We assess the performance of all models using two metrics: prediction accuracy and the area under the precision-recall curve (AUPRC). We test the proposed approach against multiple benchmarks on two widely used repositories: the UCR Repository \cite{ucr} with 128 univariate datasets and the UEA Repository \cite{uea} with 30 multivariate datasets. To further strengthen our empirical evidence, we additionally implement a K-nearest neighbor classifier equipped with DTW \cite{dtw} metric, along with T-Loss \cite{t_loss} and TST \cite{tst} beside the aforementioned SOTA approaches. 

\noindent \textbf{Results.} 
Evaluation results of our proposed CoInception framework on UCR and UEA repositories are presented in Table \ref{tab:classification}.
It is important to highlight that the results presented here pertain exclusively to 125 datasets within the UCR repository and 29 datasets in the UEA repository. The remainings are omitted to ensure fair comparisons among different baselines. 
For 125 univariate datasets in the UCR repository, CoInception ranks first in a majority of 86 datasets, and for 29 UEA datasets it produces the best classification accuracy in 16 datasets. 
In this table, we also add a detailed number of parameters for every framework when setting a fixed latent dimension of $320$. With the fusion of dilated convolution and Inception strategy, CoInception achieves the best performance while being much more lightweight ($2.35$ times) than the second best framework \cite{ts2vec}.
We also visualize the critical difference diagram \cite{cd_rank} for the Nemenyi tests on 125 UCR datasets in Figure \ref{fig:critical_diag}. Intuitively, in this diagram, classifiers connected by a bold line indicate a statistically insignificant difference in average ranks. As suggested, CoInception makes a clear improvement gap compared with other SOTAs in average ranks. 

\begin{table*}[ht]
\begin{minipage}{0.48\linewidth}
\setlength\doublerulesep{0.7pt}
\centering
\caption{Ablation analysis for the 
 proposed CoInception framework.}
\label{tab:ablation}
\centering
\scriptsize
\setlength\tabcolsep{3pt} 
\resizebox{\columnwidth}{!}{%
\begin{tabular}{lcccc}
\toprule[1pt]\midrule[0.3pt]
     & CoInception (1)                         & CoInception (2)  & CoInception (3)   & \textcolor{red}{\textbf{CoInception}} \\ \midrule
\multicolumn{5}{l}{\textit{Classification:}}                                                                                        \\
Acc. & 0.645 \textbf{(- 8.51\%)} & \textcolor{blue}{0.661 \textbf{(- 6.24\%)}} & 0.624 \textbf{(- 11.48\%}) & \textcolor{red}{0.705}                                       \\
AUC. & 0.704 \textbf{(- 9.04\%)}                        & \textcolor{blue}{0.726 \textbf{(- 6.20\%)}} & 0.691 \textbf{(- 10.72\%)} & \textcolor{red}{0.774}                                       \\ \midrule
\multicolumn{5}{l}{\textit{Forecasting:}}                                                                                           \\
MSE  & 0.067 \textbf{(- 8.95\%)}                        & 0.065 \textbf{(- 6.15\%)} & \textcolor{blue}{0.064 \textbf{(- 4.68\%)}}  & \textcolor{red}{0.061}                                     \\
MAE  & 0.178 \textbf{(- 2.81\%)}                        & 0.180 \textbf{(- 3.88\%)} & \textcolor{blue}{0.177 \textbf{(- 2.26\%)}}  & \textcolor{red}{0.173}                                       \\ \midrule
\multicolumn{5}{l}{\textit{Anomaly Detection:}}                                                                                     \\
F1   & 0.646 \textbf{(- 15.99\%)}                       & \textcolor{blue}{0.704 \textbf{(- 8.45\%)}} & 0.636 \textbf{(-17.29\%)}  & \textcolor{red}{0.769}                                       \\
P.   & 0.607 \textbf{(- 23.16\%)}                       & \textcolor{blue}{0.720 \textbf{(- 8.86\%)}} & 0.581 \textbf{(-26.45\%)}  & \textcolor{red}{0.790}                                       \\
R.   & 0.692 \textbf{(- 7.48\%)}                        & 0.689 \textbf{(- 7.88\%)} & \textcolor{blue}{0.701 \textbf{(- 6.28\%)}}  & \textcolor{red}{0.748}                                       \\ \midrule[0.3pt]\bottomrule[1pt]
\end{tabular}
}
\end{minipage}
\hfill
\begin{minipage}{0.5\linewidth}
    \begin{subfigure}{0.49\columnwidth}
    \centering
    \includegraphics[width=\textwidth]{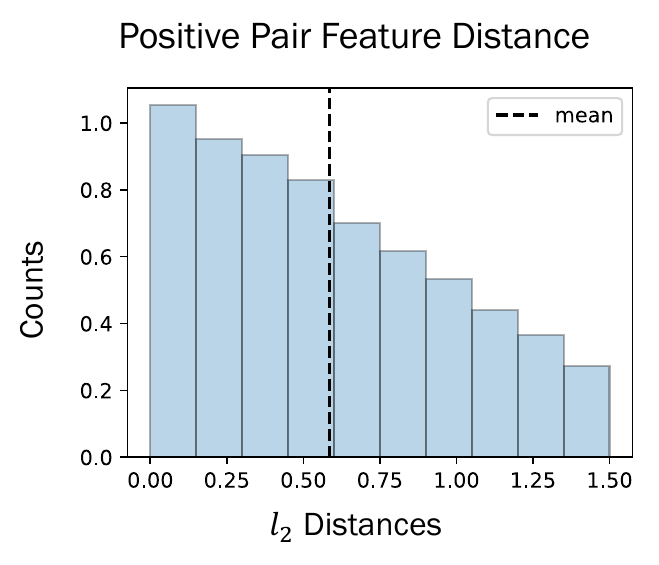}
    \caption{CoInception}
    \label{fig:coincept_align}
    \end{subfigure}
    \hfill
    \begin{subfigure}{0.49\columnwidth}
    \centering
    \includegraphics[width=\textwidth]{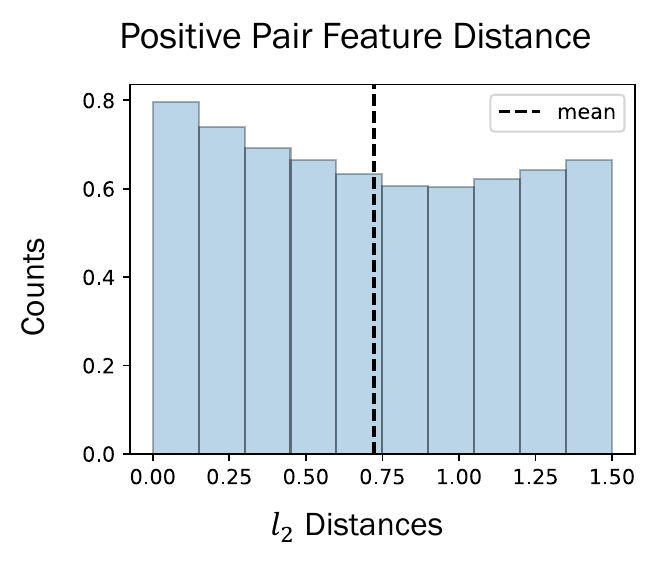}
    \caption{TS2Vec}
    \label{fig:ts2vec_align}
    \end{subfigure}
    \captionof{figure}{\textbf{Alignment analysis.} Distribution of $l_2$ distance
between features of positive pairs.}  \label{fig:align}
\end{minipage}
\end{table*}

\subsection{Time-Series Anomaly Detection}  
 
\textbf{Datasets \& Settings.} For this task, we adopt the protocols introduced by \cite{ren2019time, ts2vec}. Differently, we make three forward passes during the evaluation process to produce final prediction of CoInception. In the first pass, we mask $x^M$ and generate the corresponding representation ${z}^M_1$. The second pass puts the input segment $\mathbf{x}$ through DWT low-pass filter (Section \ref{sec:sampling_strategy}) to generate the perturbed segment $\Tilde{\mathbf{x}}$, before getting the representation ${z}^M_2$. The normal input is used in the last pass, and ${z}^M_3$ is its corresponding output. Accordingly, we define the abnormal score as $\alpha_M = \frac{1}{2}\left( \lVert z^M_1 - z^M_3 \rVert_1 + \lVert z^M_2 - z^M_3 \rVert_1 \right)$.
We keep the remaining settings intact as \cite{ren2019time, ts2vec} for both normal and cold-start experiments. Precision (P), Recall (R), and F1 score (F1) are used to evaluate anomaly detection performance.
We use the Yahoo dataset \cite{yahoo} and the KPI dataset \cite{ren2019time} from the AIOPS Challenge.
Additionally, we compare CoInception with other SOTA unsupervised methods that are utilized for detecting anomalies, such as SPOT \cite{spot}, DSPOT \cite{spot} and SR \cite{ren2019time} for normal detection tasks, as well as FFT \cite{fft} and Twitter-AD \cite{twitter} for cold-start detection tasks that require no training data.


\noindent \textbf{Results.} 
Table \ref{tab:abnormally_detection} presents a performance comparison of various methods on the Yahoo and KPI datasets using F1 score, precision, and recall metrics. We observe that CoInception outperforms existing SOTAs in the main F1 score 
for all two datasets in both the normal setting and the cold-start setting. 
In addition, CoInception also reveals its ability to perform transfer learning from one dataset to another, through steady enhancements in the empirical result for cold-start settings. 
This transferability characteristic is potentially a key to attaining a general framework for time series data. 

\section{Analysis}
 In all analyses, we sample a subset of datasets used in main experiments, which still encompasses all three main tasks to have an overall view of the performance. Regarding the classification task, we present average performance metrics across a set of 5 UCR datasets and 5 UEA datasets. 5 datasets in UCR repository include \emph{Rock, PigCVP, CinCECGTorso, SemgHandMovementCh2, HouseTwenty}; while the chosen datasets from UEA repository are \emph{DuckDuckGeese, AtrialFibrillation, Handwriting, RacketSports, SelfRegulationSCP1}. In the context of the forecasting task, we execute univariate experiments utilizing the ETTm1 dataset, with the results averaged across various prediction horizons, encompassing both short-term and long-term forecasts. As for the anomaly detection task, we offer scores for the Yahoo dataset under normal circumstances.
 
\subsection{Ablation Analysis}
\begin{figure*}[tbh]
    \centering
    \begin{subfigure}{0.55\textwidth}
    \centering
    \includegraphics[width=\textwidth]{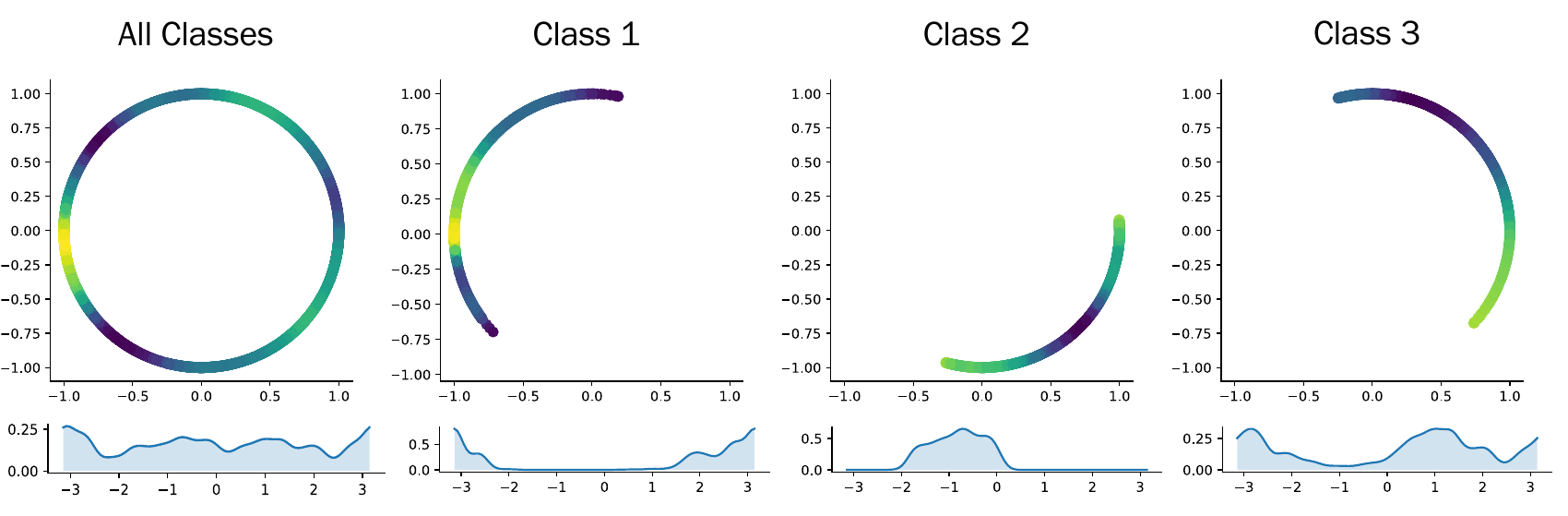}
    \caption{CoInception}
    \label{fig:coincept_unity}
    \end{subfigure}
    \begin{subfigure}{0.55\textwidth}
    \centering
    \includegraphics[width=\textwidth]{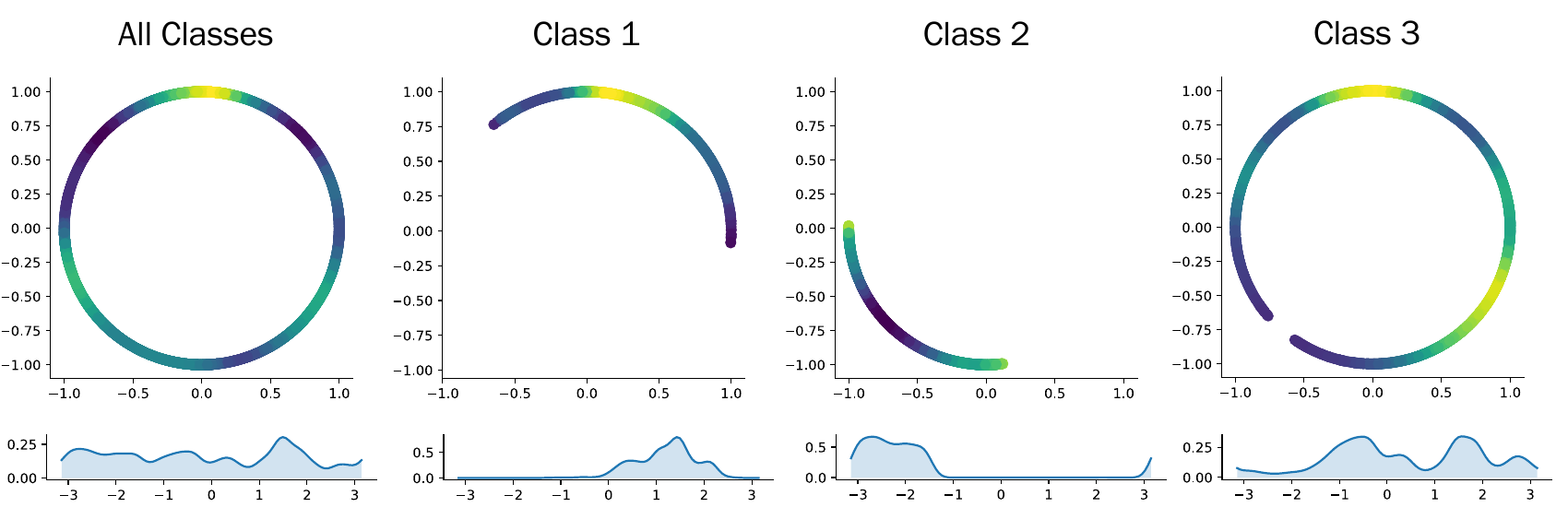}
    \caption{TS2Vec}
    \label{fig:ts2vec_unity}
    \end{subfigure}
    \caption{\textbf{Uniformity analysis.} Feature distributions with Gaussian kernel density estimation (KDE) (above) and von Mises-Fisher (vMF) KDE on angles (below). } \label{fig:uniform}
\end{figure*}
We analyze the impact of different components on the overall performance of the CoInception framework.

\noindent \textbf{Datasets \& Settings.} We designed three variations: 
\begin{itemize}
    \item[(1)] \textbf{Excluding noise-resilient sampling}, which follows the sampling strategy and hierarchical loss from \cite{ts2vec};
    \item[(2)] \textbf{Excluding Dilated Inception block}, where a stacked Dilated Convolution network is used instead of our CoInception encoder;
    \item[(3)] \textbf{Excluding triplet loss}, which omits the triplet-based term from our $\mathcal{L}_{triplet}$ calculation.
\end{itemize}

\noindent \textbf{Results.} The results are summarized in Table \ref{tab:ablation}. Overall, substantial drops in performance are observed across all three versions in the primary time series tasks. The exclusion of noise-resilient sampling led to a performance decrease from $8\%$ in classification to $15\%$ in anomaly detection. The removal of the Dilated Inception-based encoder resulted in up to $9\%$ performance decline in anomaly detection, while the elimination of triplet loss contributed to performance reductions ranging from $4\%$ to $17\%$. 

\subsection{Alignment and Uniformity}
\label{sec:align_unity}
To comprehensively evaluate the learned representations, we assessed two fundamental qualities: Alignment and Uniformity, as introduced in \cite{align_uniform}. Alignment measures the similarity of features across samples, implying that the features of a positive pair should be robust to noise. Uniformity, on the other hand, assumes that an effectively learned feature distribution should preserve maximum information. Specifically, a well-designed feature distribution should minimize the intra-similarities of positive pairs and maximize the inter-distances of negative pairs, while maintaining a uniform feature distribution to retain information.

\noindent \textbf{Settings.} To assess Alignment, we visualize the histograms that roughly indicate the distance distribution of positive pairs. We use $L2$-norm distances for this purpose. In addition, we follow the process outlined in \cite{align_uniform} to visually assess Uniformity. The learned representations are projected into $\mathbb{R}^2$ using t-SNE \cite{t_sne}, and the resulting feature distributions are visualized using Gaussian kernel density estimation (KDE) in combination with von Mises-Fisher (vMF) KDE for angles (here $\texttt{arctan2(y;x)}$).

\noindent \textbf{Results.} Fig. \ref{fig:align} summarizes the alignment of testing set features for the StarLightCurves dataset generated by CoInception and TS2Vec \cite{ts2vec}. Generally, CoInception's features exhibit a more closely clustered distribution for positive pairs. CoInception has smaller mean distances and decreasing bin heights as distance increases, unlike TS2Vec.
\noindent As suggested by figure \ref{fig:uniform}, CoInception demonstrates superior uniform characteristics for the entire test set representation, as well as better clustering between classes. Representations of different classes reside on different segments of the unit circle.


\subsection{Noise Ratio Analysis}
\label{sec:noise_ratio}
\renewcommand{\arraystretch}{1.0}
\begin{figure*}[ht]
\begin{minipage}{0.25\linewidth}
\setlength\doublerulesep{0.8pt}
\centering
\caption{CoInception and TS2Vec performance with different noise ratio in ETTm1 dataset.
}
\label{tab:noise_ratio}
\centering
\setlength\tabcolsep{3pt} 
\resizebox{\columnwidth}{!}{%
\begin{tabular}{llcc}
\toprule[1pt]\midrule[0.3pt]
\multicolumn{2}{l}{Noise Ratio} & \multicolumn{1}{l}{CoInception} & TS2Vec           \\ \midrule
\multicolumn{1}{r}{0\%}   & MSE & \textcolor{red}{0.061}    & 0.069 (-11.59\%) \\
                          & MAE & \textcolor{red}{0.173}    & 0.186 (-6.98\%)  \\
\multicolumn{1}{r}{10\%}  & MSE & \textcolor{red}{0.17}     & 0.203 (-16.25\%) \\
                          & MAE & \textcolor{red}{0.332}    & 0.364 (-8.79\%)  \\
\multicolumn{1}{r}{20\%}  & MSE & \textcolor{red}{0.175}    & 0.209 (-4.79\%)  \\
                          & MAE & \textcolor{red}{0.336}    & 0.369 (-8.94\%)  \\
\multicolumn{1}{r}{30\%}  & MSE & \textcolor{red}{0.177}    & 0.21 (-15.71\%)  \\
                          & MAE & \textcolor{red}{0.339}    & 0.37 (-8.27\%)   \\
\multicolumn{1}{r}{40\%}  & MSE & \textcolor{red}{0.18}     & 0.211 (-14.69\%) \\
                          & MAE & \textcolor{red}{0.342}    & 0.371 (-7.81\%)  \\
\multicolumn{1}{r}{50\%}  & MSE & \textcolor{red}{0.181}    & 0.213 (-15.02\%) \\
                          & MAE & \textcolor{red}{0.343}    & 0.371 (-7.54\%)  \\ \midrule[0.3pt]\bottomrule[1pt]
\end{tabular}
}
\end{minipage}
\hfill
\begin{minipage}{0.73\linewidth}
    \centering
    \includegraphics[width=0.65\textwidth]{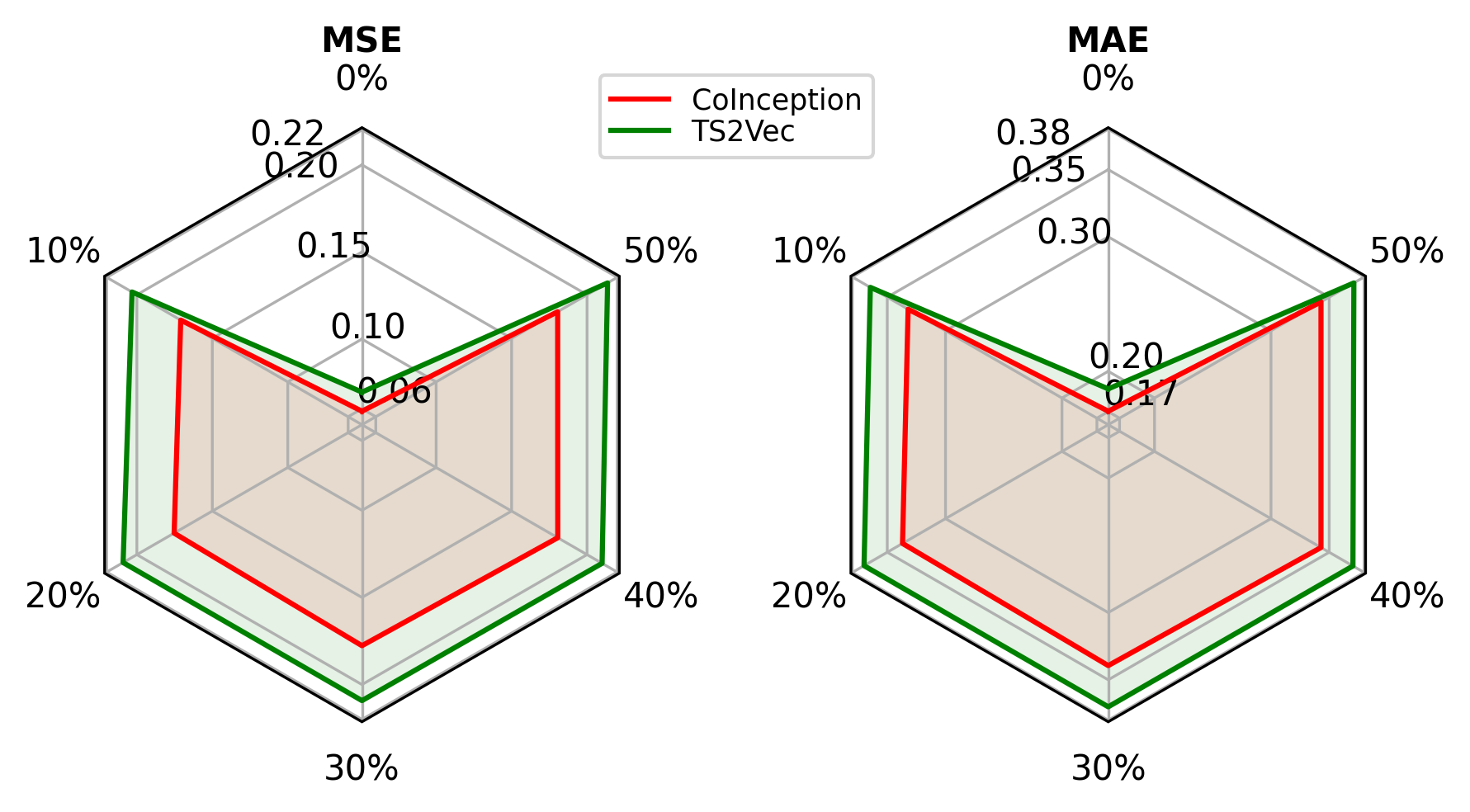}
    \caption{Assessing CoInception and TS2Vec performance with exposure to different noise ratio in ETTm1 dataset.}
    \label{fig:noise_ratio}
\end{minipage}
\end{figure*}
This experiment aims to assess the robustness of the CoInception framework under various noise ratios within a given dataset. Additionally, it aims to demonstrate the partial enhancement of noise resilience achieved by CoInception, particularly through its focus on the high-frequency component. 

\noindent \textbf{Datasets \& Settings.} For comparison, we also verify this characteristic of TS2Vec \cite{ts2vec}. 
For this experiment, we select forecasting as the representative task, using the ETTm1 dataset. By introducing random Gaussian noises with a mean equal to $x\%$ of the input series's mean attitude in pretraining stage, the goal is for two models to learn efficient representations even with the presence of noise. The current experiment sets $x$ to be 10, 20, 30, 40, and 50, as going beyond these values would result in the noise outweighing the underlying series, making it impractical to be considered as noise. We also report the results without noise (noted as $x = 0$) for complete reference. It is understandable that the model performance deteriorates when the noise level increases.

\noindent \textbf{Results.}
Figure \ref{fig:noise_ratio} and the quantitative results in Table \ref{tab:noise_ratio} summarize our findings with this experiment. In general, while both methods illustrate the decrease in performance upon the introduction of noise, CoInception still consistently outperforms TS2Vec, suggested by the performance decrease (in percentage) of TS2Vec compared with CoInception in Table \ref{tab:noise_ratio}. This results attributes to our strategy to ensure noise-resilience toward high-frequency noisy components.

\subsection{Data Augmentation with CoInception}
\label{sec:augment}
With access to label information, recent supervised frameworks \cite{patchtst, timesnet} currently demonstrate state-of-the-art performance across various time series tasks, surpassing unsupervised pipelines. This analysis effectively evaluates the potential of CoInception as a data augmentation technique, which can then be utilized by supervised frameworks to achieve new state-of-the-art results. 

\noindent \textbf{Datasets \& Settings.} We integrate CoInception as a data augmentation technique by first transforming the original datasets before feeding them into the current state-of-the-art framework, TimesNet \cite{timesnet}. Specifically, for each dataset, we pre-train CoInception using our unsupervised pipeline and generate the latent version of the corresponding datasets offline. Subsequently, for the forecasting task, this latent version is concatenated with the original dataset to form the training set for TimesNet. For the remaining two tasks, the latent datasets generated by CoInception directly replace the original ones in TimesNet's training procedure. This difference in approach stems from the labeling nature of these tasks: while forecasting utilizes the same shifted data as labels, the remaining two tasks have distinct label information.
We compare the performance of normal training strategy of vanilla TimesNet with our pipeline involving CoInception, denoted as \emph{TimesNet+CoInception}. 
\begin{table}[h]
\setlength\doublerulesep{0.8pt}
\centering
\caption{Analysis of leveraging CoInception as a data augmentation technique.}
\label{tab:augmentation}
\centering
\setlength\tabcolsep{3pt} 
\resizebox{\columnwidth}{!}{%
\begin{tabular}{@{}llllllll@{}}
\toprule[1pt]\midrule[0.3pt]
Task                                                             & \multicolumn{2}{l}{Classification}                                            & \multicolumn{2}{l}{Forecasting}                                               & \multicolumn{3}{l}{Anomaly Detection}                                                                                 \\ 
\cmidrule(lr){2-3} \cmidrule(lr){4-5} \cmidrule(lr){6-8}
                                                                 & Acc.                                  & AUC.                                  & MSE                                   & MAE                                   & F1                                    & P.                                    & R.                                    \\ 
  \midrule
TimesNet                                                         & 0.565                                 & 0.621                                 & \textcolor{red}{\textbf{0.04}}                                 & 0.147                                 & 0.374                                 & 0.519                                 & 0.292                                 \\
\begin{tabular}[c]{@{}l@{}}TimesNet +\\ CoInception\end{tabular} & \textcolor{red}{\textbf{0.592}} & \textcolor{red}{\textbf{0.664}} & 0.042 & \textcolor{red}{\textbf{0.145}} & \textcolor{red}{\textbf{0.403}} & \textcolor{red}{\textbf{0.534}} & \textcolor{red}{\textbf{0.323}} \\ \midrule[0.3pt]\bottomrule[1pt]
\end{tabular}
}
\end{table}

\noindent \textbf{Results.} Table \ref{tab:abnormally_detection} highlights the potential of the CoInception framework as a data augmentor. It demonstrates that observable improvements in performance can be attained for the classification and anomaly detection tasks, while there is minimal change for the forecasting problem. These results suggest a positive impact of CoInception in producing better starting points in the training process of unsupervised frameworks.

\section{Conclusion}\label{sec_conclusion}

We introduce CoInception, a framework for robust and efficient time series representation learning. Our approach enhances noise resilience using a pipeline involving DWT low-pass filtering and triplet-based loss. By integrating Inception blocks and dilation concepts, our encoder framework balances robustness and efficiency, outperforming state-of-the-art methods across forecasting, classification, and anomaly detection tasks.
About the limitation, we do recognize several lacks of our current work. The sampling strategy based on DWT implicitly targets high-frequency noise in this study, smoothing out the signal and revealing underlying trends or slow-varying patterns in the time series. However, this strategy may not effectively manage noise-free datasets or those with predominantly low-frequency noise. Regarding the encoder architecture, while it meets criteria for efficiency and effectiveness, the optimal number of layers remains uncertain, posing a trade-off between efficiency and effectiveness. Future studies may explore fine-tuning the number of layers depending on specific tasks or datasets.

\section*{Acknowledgement}
The work of Duy A. Nguyen was supported in part by a PhD fellowship from the VinUni-Illinois Smart Health Center, VinUniversity, Hanoi, Vietnam.

\bibliographystyle{IEEEtran}
\bibliography{IEEEabrv,abbrev,reference}

\appendix
\onecolumn
    \vbox{%
    \hsize\textwidth
    \linewidth\hsize
    \vskip 0.1in
  \hrule height 4pt
  \vskip 0.25in
  \vskip -5.5pt%
  \centering
    {\LARGE\bf{Improving Time Series Encoding with Noise-Aware Self-Supervised Learning and an Efficient Encoder \\
    Appendix} \par}
      \vskip 0.29in
  \vskip -5.5pt
  \hrule height 1pt
  \vskip 0.09in%
    
  \vskip 0.2in
}
  
\section{CoInception Supplement Details}
\subsection{Sampling Strategy}
\label{sec:app_sampling}
This section cover the working of invert DWT low-pass filter \cite{dwt} to reconstruct perturbed series $\mathbf{\Tilde{x}}$. 
This process involves combining the approximation coefficients $\mathbf{\mathbf{x}^L}$ and perturbed detail coefficients $\left\{ \mathbf{\Tilde{d}}^1, \dots, \mathbf{\Tilde{d}}^L \right\}$ obtained during the decomposition process.

By utilizing the inverse low-pass and high-pass filters, the original signal can be reconstructed from these coefficients. Let $\mathbf{\Tilde{g}}$ and $\mathbf{\Tilde{h}}$ represent these low- and high-pass filters, respectively.
The mathematical operations for the DWT reconstruction filters, which recover the perturbed signal $\mathbf{\Tilde{x}}$ at level $j$ and position $n$, can be represented as follows:
\begin{equation*}
\begin{aligned}
    \mathbf{\Tilde{x}}^{j-1}[n] &= (\mathbf{\hat{x}}^j \ast \mathbf{\Tilde{g}}^j)[n] + (\mathbf{\Tilde{d}}^j \ast \mathbf{\Tilde{h}}^j)[n] \\
    &= \sum_{k} \mathbf{\hat{x}}^j[2n-k] \mathbf{\Tilde{g}}^j[k] + \sum_{k} \mathbf{\Tilde{d}}^j[2n-k] \mathbf{\Tilde{h}}^j[k], \\
\end{aligned}
\end{equation*}
where
\begin{equation*}
\begin{cases}
\mathbf{\Tilde{x}}^L &= \mathbf{x}^L \\
\mathbf{\hat{x}}^j &= \texttt{Upsampling}(\mathbf{\Tilde{x}}^j,2).
\end{cases}
\end{equation*}
In these equations, $\mathbf{\hat{x}}^j$ represents the upsampled approximation at level $j$. 
The upsampling process \texttt{Upsampling} involves inserting zeros between the consecutive coefficients to increase their length, effectively expanding the signal toward the original length of input series. Following, the upsampled coefficients are convolved with the corresponding reconstruction filters $\mathbf{\Tilde{g}}^j$ and $\mathbf{\Tilde{h}}^j$ to obtain the reconstructed signal $\mathbf{\Tilde{x}}^{j-1}$ at the previous level.

This recursive filtering and upsampling process is repeated until the maximum useful level of decomposition, $L = \lfloor \log_2(\tfrac{M}{K}) \rfloor$, is reached. Here, $M$ represents the length of the original signal and $K$ is the length of the mother wavelet. By iteratively applying the reconstruction filters and combining the coefficients obtained after upsampling, the original signal can be reconstructed, gradually restoring both the overall trend (approximation) and the high-frequency details captured by the DWT decomposition.

\begin{figure}[ht]
    \begin{minipage}{0.48\linewidth}
        \centering
        \includegraphics[width=0.9\columnwidth]{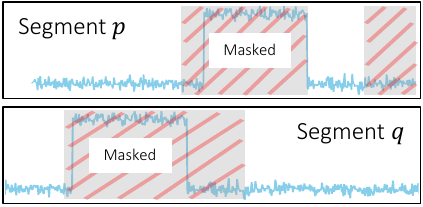}
        \caption{\textbf{Temporal Masking collapse}. An illustration for a case in which temporal masking would hinder training progress. } \label{fig:overlap}
    \end{minipage}
    \hfill
    \begin{minipage}{0.45\linewidth}
        \centering
        \includegraphics[width=\columnwidth]{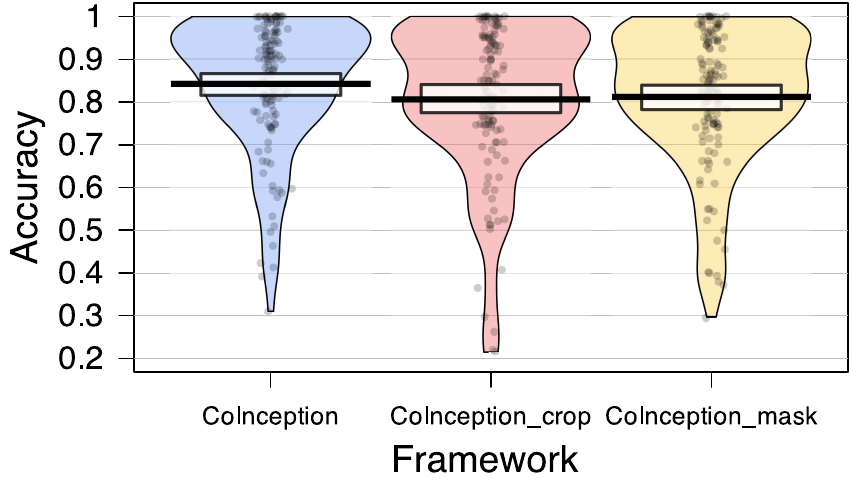}
        \caption{\textbf{Ablation in sampling strategy}. Accuracy distribution of different variants over 128 UCR dataset. } \label{fig:ablation_sampling}
    \end{minipage}
\end{figure}
To maintain the characteristic of \emph{context invariance}, we employ a variation of the approach proposed with TS2Vec \cite{ts2vec}. Specifically, we choose to rely solely on random cropping for generating overlapping segments, without incorporating temporal masking. This decision is based on recognizing several scenarios that could undermine the effectiveness of this strategy. Firstly, if heavy masking is applied, it may lead to a lack of explicit context information. The remaining context information, after extensive occlusion, might be insufficient or unrepresentative for recovering the masked timestamp, thus impeding the learning process. Secondly, when dealing with data containing occasional abnormal timestamps (e.g., level shifts), masking these timestamps in both overlapping segments (Figure \ref{fig:overlap}) can also hinder the learning progress since the contextual information becomes non-representative for inference.

According to the findings discussed in \cite{ts2vec}, random cropping is instrumental in producing position-agnostic representations, which helps prevent the occurrence of \emph{representation collapse} when using temporal contrasting. This is attributed to the inherent capability of Convolutional networks to encode positional information in their learned representations \cite{conv_1, conv_2}, thereby mitigating the impact of temporal contrasting as a learning strategy. As the Inception block in CoInception primarily consists of Convolutional layers, the adoption of random cropping assumes utmost importance in enabling CoInception to generate meaningful representations.

The accuracy distribution for 128 UCR datasets across various CoInception variations is illustrated in Figure \ref{fig:ablation_sampling}. These variations include: (1) the ablation of Random Cropping, where two similar segments are used instead, and (2) the inclusion of temporal masking on the latent representations, following the approach in \cite{ts2vec}. As depicted in the figure, both variations exhibit a decrease in overall performance and a higher variance in accuracy across the 128 UCR datasets, compared with our proposed framework.

\subsection{Inception-Based Dilated Convolution Encoder}
While the main structure of our CoInception Encoder is a stack of Inception blocks, there are some additional details discussed in this section. 

Before being fed into the first Inception block, the input segments are first projected into a different latent space, other than the original feature space.
We intentionally perform the mapping with a simple Fully Connected layer.
\begin{equation*}
\begin{split}
    &\theta: \mathbb{R}^{M \times N} \rightarrow \mathbb{R}^{M \times K} \\
    &\theta(\mathbf{x}) = \mathbf{W}\mathbf{x} + \mathbf{b}
\end{split}
\end{equation*}
The benefits of this layer are twofold. First, upon dealing with high-dimensional series, this layer essentially act as a filter for dimensionality reduction. The latent space representation retains the most informative features of the input segment while discarding irrelevant or redundant information, reducing the computational burden on the subsequent Inception blocks. This layer make CoInception more versatile to different datasets, and ensure its scalability. Second, the projection by Fully Connected layer help CoInception enhance its transferability. Upon adapting the framework trained with one dataset to another, we only need to retrain the projection layer, while keeping the main stacked Inception layers intact.

To provide a clearer understanding of the architecture depicted in Inception block \ref{fig:inception_block}, we will provide a detailed interpretation. In our implementation, each Inception block consists of three Basic units. Let the outputs of these units be denoted as $\mathbf{b}_1$, $\mathbf{b}_2$, and $\mathbf{b}_3$. To enhance comprehension, we will use the notation $\mathbf{b}_j$ to represent these three outputs collectively. Additionally, we will use $\mathbf{m}$ to denote the output of the \texttt{Maxpooling} unit, and $\mathbf{h}$ to represent the overall output of the entire Inception block. The following formulas outline the operations within the $i^{th}$ Inception block.
\begin{equation}
\begin{aligned}
    \mathbf{b}_k^i &= \texttt{Conv1d*}(\sigma(\texttt{Conv1d*}(\mathbf{h}^{i-1}))) + \sigma(\texttt{Conv1d*}(\mathbf{b}^{i-1}_k)), \\
    \mathbf{m}^i &= \sigma(\texttt{Conv1d*}(\mathbf{b}^{i-1}_k)), \\
    \mathbf{h}^i &= \texttt{Aggregator}(\texttt{Concat}(\mathbf{b}_k^i, \mathbf{m}^i)). 
\end{aligned}
\end{equation}
In these equations, $\sigma$ represent the \emph{LeakyReLU} activation function, which is used throughout the CoInception architecture.



\section{Implementation Details}
\subsection{Environment Settings}
All implementations and experiments are performed on a single machine with the following hardware configuration: an $64-$core Intel Xeon CPU with a GeForce RTX 3090 GPU to accelerate training. Our codebase primarily relies on the \emph{PyTorch 2.0} framework for deep learning tasks. Additionally, we utilize utilities from \emph{Scikit-learn, Pandas,} and \emph{Matplotlib} to support various functionalities in our experiments.

\subsection{CoInception's Reproduction}
\textbf{Sampling Strategy.}
In our current implementation for CoInception, we employ the Daubechies wavelet family \cite{daubechies_dwt}, known for its widespread use and suitability for a broad range of signals \cite{daubechies_dwt1, daubechies_dwt2, daubechies_dwt3}. Specifically, we utilize the Daubechies D4 wavelets as both low and high-pass filters in CoInception across all experiments, as mentioned in \cite{dwt_apply}. It is important to note, however, that our selection of the mother wavelet serves as a reference, and it is advisable to invest additional effort in choosing the optimal wavelets for specific datasets \cite{dwt_apply}. Such careful consideration may further enhance the accuracy of CoInception for specific tasks.

\textbf{Inception-Based Dilated Convolution Encoder.}
In our experiments, we incorporate three Inception blocks, each comprising three Basic units. The base kernel sizes employed in these blocks are $2$, $5$, and $8$ respectively. For non-linear transformations, we utilize the $LeakyReLU$ activation function consistently across the architecture. To ensure fair comparisons across all benchmarks, we maintain a constant latent dimension of $64$ and a final output representation size of $320$.

\textbf{Hierarchical Triplet Loss.}
In the calculation of $\mathcal{L}_{triplet}$ (Eq. \ref{eq:loss}), several hyperparameters are utilized. The balance factor $\epsilon$ is assigned a value of $0.7$, indicating a higher weight distribution towards minimizing the distance between positive samples. The triplet term serves as an additional constraint and receives a relatively smaller weight. For the triplet term itself, the margin $\eta$ is set to $1$.

\subsection{Baselines' Reproduction}
Due to the extensive comparison of CoInception with numerous baselines, many of which are specifically designed for particular tasks, we have chosen to reproduce results for a selected subset while inheriting results from other relevant works. Specifically, we reproduce the results from three works that focus on various time series tasks, namely TS2Vec \cite{ts2vec}, TS-TCC \cite{ts_tcc}, and TNC \cite{tnc}. The majority of the remaining results are directly sourced from \cite{ts2vec}, \cite{t_loss}, \cite{informer}, \cite{ren2019time}, and \cite{btsf}.

\subsection{Shared settings in most analyses}
\label{sec:app_setting}
As mentioned in main text, we sample a subset of datasets used in main experiments, which still encompasses all three main tasks to have an overall view of the performance. Regarding the classification task, we present average performance metrics across a set of 5 UCR datasets and 5 UEA datasets. 5 datasets in UCR repository include \emph{Rock, PigCVP, CinCECGTorso, SemgHandMovementCh2, HouseTwenty}; while the chosen datasets from UEA repository are \emph{DuckDuckGeese, AtrialFibrillation, Handwriting, RacketSports, SelfRegulationSCP1}. In the context of the forecasting task, we execute univariate experiments utilizing the ETTm1 dataset, with the results averaged across various prediction horizons, encompassing both short-term and long-term forecasts. As for the anomaly detection task, we offer scores for the Yahoo dataset under normal circumstances.

\section{Further Experiment Results and Analysis}
\subsection{Time Series Forecasting}
\label{sec:app_forecast}
\textbf{Additional details.} During the data processing stage, z-score normalization is applied to each feature in both the univariate and multivariate datasets. All reported results are based on scores obtained from these normalized datasets.
In the univariate scenario, additional features are introduced alongside the main feature, following a similar approach as described in \cite{informer, ts2vec}. These additional features include \textit{minute}, \textit{hour}, \textit{day of week}, \textit{day of month}, \textit{day of year}, \textit{month of year}, and \textit{week of year}.
For the train-test split, the first 12 months are used for training, followed by 4 months for validation, and the last 4 months for three ETT datasets, following the methodology outlined in \cite{informer}. In the case of the Electricity dataset, a ratio of $60-20-20$ is used for the train, validation, and test sets, respectively, following \cite{ts2vec}.

After the completion of the unsupervised training phase, the learned representations are evaluated using a forecasting task, following a protocol similar to \cite{ts2vec}. A linear regression model with an $L_2$ regularization term $\alpha$ is employed. The value of $\alpha$ is chosen through a grid search over the search space $\left\{0.1, 0.2, 0.5, 1, 2, 5, 10, 20, 50, 100, 200, 500, 1000\right\}$.

\textbf{Additional results.} The full results for the univariate and multivariate forecasting experiments are presented in Table \ref{tab:univar_forecasting} and Table \ref{tab:mulvar_forecasting}, respectively. For both circumstances, CoInception demonstrates its superiority in every testing dataset, in most configurations for the output number of forecasting timesteps (highlighted with \textcolor{red}{\textbf{bold, red numbers}}). 
\begin{table*}[ht]
\setlength\doublerulesep{0.8pt}
\centering
\caption{Univariate time series forecasting results. Best results are bold and highlighted in \textcolor{red}{\textbf{red}}, and second best results are in \textcolor{blue}{blue}.}
\label{tab:univar_forecasting}
\centering
\small
\setlength\tabcolsep{3pt} 
\begin{tabular}{lccccccccccccccc}
\toprule[1pt]\midrule[0.3pt]
                              &                     & \multicolumn{2}{c}{TS2Vec}                                  & \multicolumn{2}{c}{TS-TCC} & \multicolumn{2}{c}{TNC}              & \multicolumn{2}{c}{Informer}                                & \multicolumn{2}{c}{N-BEATS} & \multicolumn{2}{c}{TimesNet*} & \multicolumn{2}{c}{CoInception}                                      \\  \cmidrule(lr){3-4} \cmidrule(lr){5-6} \cmidrule(lr){7-8} \cmidrule(lr){9-10} \cmidrule(lr){11-12} \cmidrule(lr){13-14} \cmidrule(lr){15-16} 
\multirow{-2}{*}{Dataset}     & \multirow{-2}{*}{T} & MSE                                   & MAE                                   & MSE                                   & MAE                          & MSE   & MAE                          & MSE                                   & MAE                                   & MSE          & MAE          & MSE           & MAE          & MSE                                   & MAE                                   \\ \midrule
                              & 24                  & \textcolor{red}{ \textbf{0.039}} & \textcolor{red}{ \textbf{0.152}} & 0.117                                 & 0.281                        & 0.075 & 0.21                         & 0.098                                 & 0.247                                 & 0.094        & 0.238        & 0.091         & 0.233        & \textcolor{red}{ \textbf{0.039}} & \textcolor{blue}{0.153}          \\
                              & 48                  & \textcolor{red}{ \textbf{0.062}} & \textcolor{red}{ \textbf{0.191}} & 0.192                                 & 0.369                        & 0.227 & 0.402                        & 0.158                                 & 0.319                                 & 0.21         & 0.367        & 0.154         & 0.307        & \textcolor{blue}{0.064}          & \textcolor{blue}{0.196}          \\
                              & 168                 & \textcolor{blue}{0.134}          & \textcolor{blue}{0.282}          & 0.331                                 & 0.505                        & 0.316 & 0.493                        & 0.183                                 & 0.346                                 & 0.232        & 0.391        & 0.293         & 0.336        & \textcolor{red}{ \textbf{0.128}} & \textcolor{red}{ \textbf{0.275}} \\
                              & 336                 & \textcolor{blue}{0.154}          & \textcolor{blue}{0.31}           & 0.353                                 & 0.525                        & 0.306 & 0.495                        & 0.222                                 & 0.387                                 & 0.232        & 0.388        & 0.334          & 0.417        & \textcolor{red}{ \textbf{0.15}}  & \textcolor{red}{ \textbf{0.303}} \\
\multirow{-5}{*}{ETTh1}       & 720                 & \textcolor{blue}{0.163}          & \textcolor{blue}{0.327}          & 0.387                                 & 0.560                        & 0.39  & 0.557                        & 0.269                                 & 0.435                                 & 0.322        & 0.49         & 0.499         & 0.545        & \textcolor{red}{ \textbf{0.161}} & \textcolor{red}{ \textbf{0.317}} \\ \midrule
                              & 24                  & \textcolor{blue}{0.090}          & \textcolor{blue}{0.229}          & 0.106                                 & 0.255                        & 0.103 & 0.249                        & 0.093                                 & 0.240                                 & 0.198        & 0.345        & 0.095         & 0.249        & \textcolor{red}{ \textbf{0.086}} & \textcolor{red}{ \textbf{0.217}} \\
                              & 48                  & \textcolor{blue}{0.124}          & \textcolor{blue}{0.273}          & 0.138                                 & 0.293                        & 0.142 & 0.290                        & 0.155                                 & 0.314                                 & 0.234        & 0.386        & 0.252         & 0.31        & \textcolor{red}{ \textbf{0.119}} & \textcolor{red}{ \textbf{0.264}} \\
                              & 168                 & \textcolor{blue}{0.208}          & \textcolor{blue}{0.360}          & 0.211                                 & 0.368                        & 0.227 & 0.376                        & 0.232                                 & 0.389                                 & 0.331        & 0.453        & 0.368         & 0.381        & \textcolor{red}{ \textbf{0.185}} & \textcolor{red}{ \textbf{0.339}} \\
                              & 336                 & \textcolor{blue}{0.213}          & \textcolor{blue}{0.369}          & 0.222                                 & 0.379                        & 0.296 & 0.430                        & 0.263                                 & 0.417                                 & 0.431        & 0.508        & 0.459         & 0.513        & \textcolor{red}{ \textbf{0.196}} & \textcolor{red}{ \textbf{0.353}} \\
\multirow{-5}{*}{ETTh2}       & 720                 & \textcolor{blue}{0.214}          & \textcolor{blue}{0.374}          & 0.238                                 & 0.394                        & 0.325 & 0.463                        & 0.277                                 & 0.431                                 & 0.437        & 0.517        & 0.561         & 0.612        & \textcolor{red}{ \textbf{0.209}} & \textcolor{red}{ \textbf{0.370}} \\ \midrule
                              & 24                  & \textcolor{blue}{0.015}          & \textcolor{blue}{0.092}          & 0.048                                 & 0.172                        & 0.041 & 0.157                        & 0.03                                  & 0.137                                 & 0.054        & 0.184        & 0.105         & 0.237        & \textcolor{red}{ \textbf{0.013}} & \textcolor{red}{ \textbf{0.083}} \\
                              & 48                  & \textcolor{blue}{0.027}          & \textcolor{blue}{0.126}          & 0.076                                 & 0.219                        & 0.101 & 0.257                        & 0.069                                 & 0.203                                 & 0.190        & 0.361        & 0.152         & 0.279        & \textcolor{red}{ \textbf{0.025}} & \textcolor{red}{ \textbf{0.116}} \\
                              & 96                  & \textcolor{blue}{0.044}          & \textcolor{blue}{0.161}          & 0.116                                 & 0.277                        & 0.142 & 0.311                        & 0.194                                 & 0.372                                 & 0.183        & 0.353        & 0.158         & 0.295        & \textcolor{red}{ \textbf{0.041}} & \textcolor{red}{ \textbf{0.152}} \\
                              & 288                 & \textcolor{blue}{0.103}          & \textcolor{blue}{0.246}          & 0.233                                 & 0.413                        & 0.318 & 0.472                        & 0.401                                 & 0.554                                 & 0.186        & 0.362        & 0.286         & 0.339        & \textcolor{red}{ \textbf{0.092}} & \textcolor{red}{ \textbf{0.231}} \\
\multirow{-5}{*}{ETTm1}       & 672                 & \textcolor{blue}{0.156}          & \textcolor{blue}{0.307}          & 0.344                                 & 0.517                        & 0.397 & 0.547                        & 0.512                                 & 0.644                                 & 0.197        & 0.368        & 0.292         & 0.348        & \textcolor{red}{ \textbf{0.138}} & \textcolor{red}{ \textbf{0.287}} \\ \midrule
                              & 24                  & 0.260                                 & 0.288                                 & 0.261                                 & 0.297                        & 0.263 & \textcolor{blue}{0.279} & \textcolor{red}{ \textbf{0.251}} & \textcolor{red}{ \textbf{0.275}} & 0.427        & 0.330        & 0.408        & 0.417        & \textcolor{blue}{0.256}          & 0.288                                 \\
                              & 48                  & \textcolor{blue}{0.319}          & \textcolor{blue}{0.324}          & \textcolor{red}{ \textbf{0.307}} & \textcolor{blue}{0.319} & 0.373 & 0.344                        & 0.346                                 & 0.339                                 & 0.551        & 0.392        & 0.578         & 0.529        & \textcolor{red}{ \textbf{0.307}}          & \textcolor{red}{ \textbf{0.317}} \\
                              & 168                 & \textcolor{blue}{0.427}          & \textcolor{blue}{0.394}          & 0.438                                 & 0.403                        & 0.609 & 0.462                        & 0.544                                 & 0.424                                 & 0.893        & 0.538        & 0.955         & 0.718        & \textcolor{red}{ \textbf{0.426}} & \textcolor{red}{ \textbf{0.391}} \\
                              & 336                 & \textcolor{blue}{0.565}          & \textcolor{blue}{0.474}          & 0.592                                 & 0.478                        & 0.855 & 0.606                        & 0.713                                 & 0.512                                 & 1.035        & 0.669        & 1.016         & 0.806        & \textcolor{red}{ \textbf{0.56}}  & \textcolor{red}{ \textbf{0.472}} \\
\multirow{-5}{*}{Elec.} & 720                 & \textcolor{blue}{0.861}          & \textcolor{blue}{0.643}          & 0.885                                 & 0.663                        & 1.263 & 0.858                        & 1.182                                 & 0.806                                 & 1.548        & 0.881        & 1.105         & 0.819        & \textcolor{red}{ \textbf{0.859}} & \textcolor{red}{ \textbf{0.638}} \\ \midrule[0.3pt]\bottomrule[1pt]
\end{tabular}
\end{table*}
\renewcommand{\arraystretch}{1.0}
\begin{table*}[ht]
\setlength\doublerulesep{0.8pt}
\centering
\caption{Multivariate time series forecasting results. 
}
\label{tab:mulvar_forecasting}
\centering
\small
\setlength\tabcolsep{3pt} 
\begin{tabular}{lccccccccccccccc}
\toprule[1pt]\midrule[0.3pt]
\multicolumn{1}{c}{\multirow{2}{*}{Dataset}} & \multirow{2}{*}{T} & \multicolumn{2}{c}{TS2Vec} & \multicolumn{2}{c}{TS-TCC} & \multicolumn{2}{c}{TNC} & \multicolumn{2}{c}{Informer}    & \multicolumn{2}{c}{StemGNN} & \multicolumn{2}{c}{TimesNet*} & \multicolumn{2}{c}{\textcolor{red}{\textbf{CoInception}}} \\ \cmidrule(lr){3-4} \cmidrule(lr){5-6} \cmidrule(lr){7-8} \cmidrule(lr){9-10} \cmidrule(lr){11-12} \cmidrule(lr){13-14} \cmidrule(lr){15-16} 
\multicolumn{1}{c}{}                         &                         & MSE              & MAE     & MSE          & MAE         & MSE        & MAE        & MSE            & MAE            & MSE               & MAE     & MSE      & MAE               & MSE               & MAE               \\ \midrule
                                             & 24                       & 0.599                                 & {\color{blue} 0.534} & 0.653                        & 0.610                        & 0.632                        & 0.596                        & {\color{blue} 0.577}          & 0.549                                 & 0.614                                 & 0.571                        & 0.914                        & 0.836                                 & {\color{red} \textbf{0.461}} & {\color{red} \textbf{0.479}} \\
                                              & 48                       & {\color{blue} 0.629}          & {\color{blue} 0.555} & 0.720                        & 0.693                        & 0.705                        & 0.688                        & 0.685                                 & 0.625                                 & 0.748                                 & 0.618                        & 1.006                        & 0.971                                 & {\color{red} \textbf{0.512}} & {\color{red} \textbf{0.503}} \\
                                              & 168                      & 0.755                                 & 0.636                        & 1.129                        & 1.044                        & 1.097                        & 0.993                        & 0.931                                 & 0.752                                 & {\color{red} \textbf{0.663}} & {\color{blue} 0.608} & 1.105                        & 1.066                                 & {\color{blue} 0.683}          & {\color{red} \textbf{0.601}} \\
                                              & 336                      & {\color{blue} 0.907}          & {\color{blue} 0.717} & 1.492                        & 1.076                        & 1.454                        & 0.919                        & 1.128                                 & 0.873                                 & 0.927                                 & 0.730                        & 1.15                        & 1.097                                 & {\color{red} \textbf{0.829}} & {\color{red} \textbf{0.678}} \\
\multirow{-5}{*}{ETTh1}                       & 720                      & {\color{blue} 1.048}          & {\color{blue} 0.790} & 1.603                        & 1.206                        & 1.604                        & 1.118                        & 1.215                                 & 0.896                                 & -                                     & -                            & 1.348                        & 1.212                                 & {\color{red} \textbf{1.018}} & {\color{red} \textbf{0.770}} \\ \midrule
                                              & 24                       & {\color{blue} 0.398}          & {\color{blue} 0.461} & 0.883                        & 0.747                        & 0.830                        & 0.756                        & 0.720                                 & 0.665                                 & 1.292                                 & 0.883                        & 0.915                        & 0.866                                 & {\color{red} \textbf{0.335}} & {\color{red} \textbf{0.432}} \\
                                              & 48                       & {\color{blue} 0.580}          & {\color{blue} 0.573} & 1.701                        & 1.378                        & 1.689                        & 1.311                        & 1.457                                 & 1.001                                 & 1.099                                 & 0.847                        & 1.709                        & 0.981                                 & {\color{red} \textbf{0.550}} & {\color{red} \textbf{0.560}} \\
                                              & 168                      & {\color{blue} 1.901}          & {\color{blue} 1.065} & 3.956                        & 2.301                        & 3.792                        & 2.029                        & 3.489                                 & 1.515                                 & 2.282                                 & 1.228                        & 2.224                        & 1.155                                 & {\color{red} \textbf{1.812}} & {\color{red} \textbf{1.055}} \\
                                              & 336                      & {\color{blue} 2.304}          & {\color{blue} 1.215} & 3.992                        & 2.852                        & 3.516                        & 2.812                        & 2.723                                 & 1.340                                 & 3.086                                 & 1.351                        & 3.017                        & 1.359                                 & {\color{red} \textbf{2.151}} & {\color{red} \textbf{1.188}} \\
\multirow{-5}{*}{ETTh2}                       & 720                      & {\color{red} \textbf{2.650}} & {\color{blue} 1.373} & 4.732                        & 2.345                        & 4.501                        & 2.410                        & 3.467                                 & 1.473                                 & -                                     & -                            & 3.121                        & 1.466                                 & {\color{blue} 2.962}          & {\color{red} \textbf{1.338}} \\ \midrule
                                              & 24                       & 0.443                                 & 0.436                        & 0.473                        & 0.490                        & 0.429                        & 0.455                        & {\color{red} \textbf{0.323}} & {\color{red} \textbf{0.369}} & 0.620                                 & 0.570                        & 1.005                        & 0.791                                 & {\color{blue} 0.384}          & {\color{blue} 0.423}          \\
                                              & 48                       & 0.582                                 & {\color{blue} 0.515} & 0.671                        & 0.665                        & 0.623                        & 0.602                        & {\color{red} \textbf{0.494}} & {\color{red} \textbf{0.503}} & 0.744                                 & 0.628                        & 1.008                        & 0.848                                 & {\color{blue} 0.552}          & 0.521                                 \\
                                              & 96                      & {\color{blue} 0.622}          & {\color{blue} 0.549} & 0.803                        & 0.724                        & 0.749                        & 0.731                        & 0.678                                 & 0.614                                 & 0.709                                 & 0.624                        & 1.104                        & 0.916                                 & {\color{red} \textbf{0.561}} & {\color{red} \textbf{0.533}} \\
                                              & 288                      & {\color{blue} 0.709}          & {\color{blue} 0.609} & 1.958                        & 1.429                        & 1.791                        & 1.356                        & 1.056                                 & 0.786                                 & 0.843                                 & 0.683                        & 1.109                        & 0.969                                 & {\color{red} \textbf{0.623}} & {\color{red} \textbf{0.578}} \\
\multirow{-5}{*}{ETTm1}                       & 672                      & {\color{blue} 0.786}          & {\color{blue} 0.655} & 1.838                        & 1.601                        & 1.822                        & 1.692                        & 1.192                                 & 0.926                                 & -                                     & -                            & 1.115                        & 1.011                                 & {\color{red} \textbf{0.717}} & {\color{red} \textbf{0.639}} \\ \midrule
                                              & 24                       & 0.287                                 & 0.374                        & {\color{blue} 0.278} & {\color{blue} 0.370} & 0.305                        & 0.384                        & 0.312                                 & 0.387                                 & 0.439                                 & 0.388                        & 0.414                        & 0.529                                 & {\color{red} \textbf{0.234}} & {\color{red} \textbf{0.335}} \\
                                              & 48                       & {\color{blue} 0.307}          & {\color{blue} 0.388} & 0.313                        & 0.392                        & 0.317                        & 0.392                        & 0.392                                 & 0.431                                 & 0.413                                 & 0.455                        & 0.516                        & 0.63                                 & {\color{red} \textbf{0.265}} & {\color{red} \textbf{0.356}} \\
                                              & 168                      & {\color{blue} 0.332}          & {\color{blue} 0.407} & 0.338                        & 0.411                        & 0.358                        & 0.423                        & 0.515                                 & 0.509                                 & 0.506                                 & 0.518                        & 0.585                        & 0.684                                 & {\color{red} \textbf{0.282}} & {\color{red} \textbf{0.372}} \\
                                              & 336                      & {\color{blue} 0.349}          & 0.420                        & 0.357                        & 0.424                        & {\color{blue} 0.349} & {\color{blue} 0.416} & 0.759                                 & 0.625                                 & 0.647                                 & 0.596                        & 0.601                        & 0.692                                 & {\color{red} \textbf{0.301}} & {\color{red} \textbf{0.388}} \\
\multirow{-5}{*}{Elec.}                 & 720                      & {\color{blue} 0.375}                                 & {\color{blue} 0.438}                        & 0.382                        & 0.442                        & 0.447                        & 0.486                        & 0.969                                 & 0.788                                 & -                                     & -                            & 0.663 & 0.727 & {\color{red} \textbf{0.331}} & {\color{red} \textbf{0.409}}          \\
\midrule[0.3pt]\bottomrule[1pt]
\end{tabular}
\end{table*}

\subsection{Time Series Classification}
\label{sec:app_classify}
\textbf{Additional details.}
During the data processing stage, all datasets from the UCR Repository are normalized using z-score normalization, resulting in a mean of $0$ and a variance of $1$. Similarly, for datasets from the UEA Repository, each feature is independently normalized using z-score normalization. It is important to note that within the UCR Repository, there are three datasets that contain missing data points: \textit{DodgerLoopDay}, \textit{DodgerLoopGame}, and \textit{DodgerLoopWeekend}. These datasets cannot be handled with T-Loss, TS-TCC, or TNC methods. However, with the employment of CoInception, we address this issue by directly replacing the missing values with $0$ and proceed with the training process as usual.

As stated in the main manuscript, the representations generated by CoInception are passed through a \texttt{MaxPooling} layer to extract the representative timestamp, which serves as the instance-level representation of the input. This instance-level representation is subsequently utilized as the input for training the classifier. Consistent with \cite{t_loss, ts2vec}, we employ a Radial Basis Function (RBF) Support Vector Machine (SVM) classifier. The penalty parameter $C$ for the SVM is selected through a grid search conducted over the range $\left\{ 10^i | i \in [-4,4]\right\}$.

\textbf{Additional results.} The comprehensive results of our CoInception framework on the 128 UCR Datasets, along with other baselines (TS2Vec \cite{ts2vec}, T-Loss \cite{t_loss}, TS-TCC \cite{ts_tcc}, TST \cite{tst}, TNC \cite{tnc}, and DWT \cite{dtw}), are presented in Table \ref{tab:classification_ucr}. In general, CoInception outperforms other state-of-the-art methods in $67\%$ of the 128 datasets from the UCR Repository.

\begin{figure}[ht]
\begin{minipage}{0.53\linewidth}
    \centering
    \includegraphics[width=\columnwidth]{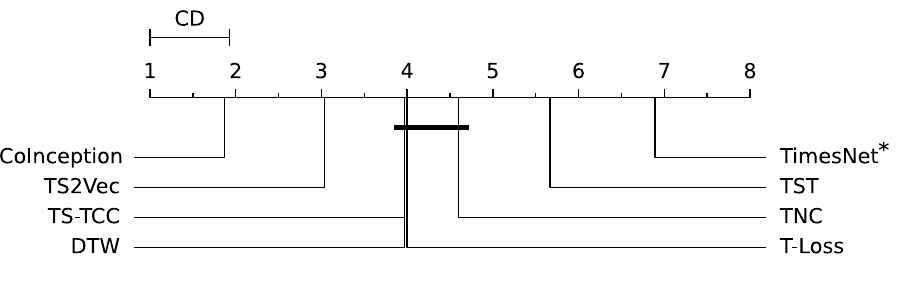}
    \caption{\textbf{Critical Difference Diagram}. Different classifiers' ranks on 29 Datasets from UEA Repository with the
confidence level of $95\%$. } \label{fig:cd_uea}

\end{minipage}
\hfill
\begin{minipage}{0.43\linewidth}
    \centering
    \includegraphics[width=\columnwidth]{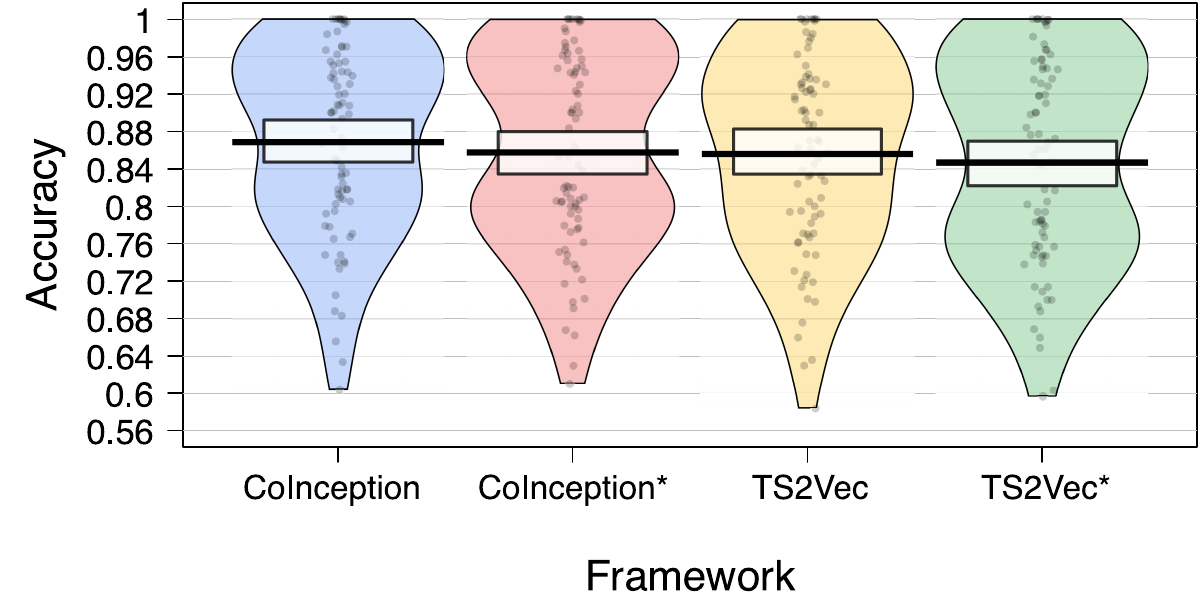}
    \caption{\textbf{Transferability Analysis}. Accuracy distribution for the first 85 UCR datasets.}
\end{minipage}
\end{figure}

Similarly, detailed results for the 30 UEA Repository datasets are summarized in Table \ref{tab:classification_uea}, accompanied by the corresponding Critical Difference diagram for the first 29 datasets depicted in Figure \ref{fig:cd_uea}. In line with the findings in the univariate setting, CoInception also achieves better performance than more than $55\%$ of the datasets in the UEA Repository's multivariate scenario.

From both tables, it is evident that CoInception exhibits superior performance for the majority of datasets, resulting in a significant performance gap in terms of average accuracy.
\begin{table*}[ht]
\setlength\doublerulesep{0.8pt}
\centering
\caption{UCR 128 Datasets classification results. Best results are bold and highlighted in \textcolor{red}{\textbf{red}}.}
\label{tab:classification_ucr}
\centering
\setlength\tabcolsep{3pt} 
\footnotesize
\begin{tabular}{lccccccccc}
\toprule[1pt]\midrule[0.3pt]
Dataset                        & TS2Vec                                & T-Loss                                & TNC                                   & TS-TCC                                & TST                                   & DTW                                  & TimesNet*                             & CoInception                           & CoInception's Rank \\ \midrule
Adiac                          & 0.762                                 & 0.675                                 & 0.726                                 & \textcolor{red}{\textbf{0.767}} & 0.55                                  & 0.604                                & 0.66                                  & \textcolor{red}{\textbf{0.767}} & 1                  \\
ArrowHead                      & 0.857                                 & 0.766                                 & 0.703                                 & 0.737                                 & 0.771                                 & 0.703                                & 0.794                                 & \textcolor{red}{\textbf{0.863}} & 1                  \\
Beef                           & \textcolor{red}{\textbf{0.767}} & 0.667                                 & 0.733                                 & 0.6                                   & 0.5                                   & 0.633                                & 0.567                                 & 0.733                                 & 2                  \\
BeetleFly                      & 0.9                                   & 0.8                                   & 0.85                                  & 0.8                                   & \textcolor{red}{\textbf{1}}     & 0.7                                  & 0.65                                  & 0.85                                  & 3                  \\
\end{tabular}
\end{table*}
\begin{table*}[ht]
\setlength\doublerulesep{0.8pt}
\centering
\centering
\footnotesize
\setlength\tabcolsep{3pt} 
\begin{tabular}{lccccccccc}
\toprule[1pt]\midrule[0.3pt]
Dataset                   & TS2Vec                                & T-Loss                                & TNC                                   & TS-TCC                                & TST                                  & DTW                                   & TimesNet*                             & CoInception                           & CoInception's Rank \\ \midrule
Strawberry                & 0.962                                 & 0.954                                 & 0.951                                 & \textcolor{red}{\textbf{0.965}} & 0.916                                & 0.941                                 & 0.916                                 & \textcolor{red}{\textbf{0.97}}  & 1                  \\
SwedishLeaf               & 0.941                                 & 0.914                                 & 0.88                                  & 0.923                                 & 0.738                                & 0.792                                 & 0.853                                 & \textcolor{red}{\textbf{0.95}}  & 1                  \\
Symbols                   & \textcolor{red}{\textbf{0.976}} & 0.963                                 & 0.885                                 & 0.916                                 & 0.786                                & 0.95                                  & 0.886                                 & 0.97                                  & 2                  \\
SyntheticControl          & 0.997                                 & 0.987                                 & \textcolor{red}{\textbf{1}}     & 0.99                                  & \textcolor{red}{\textbf{0.49}} & 0.993                                 & 0.98                                  & 0.997                                 & 2                  \\
\end{tabular}
\end{table*}
\begin{table*}[ht]
\setlength\doublerulesep{0.8pt}
\centering
\caption{UEA 30 Datasets classification results. Best results are bold and highlighted in \textcolor{red}{\textbf{red}}.}
\label{tab:classification_uea}
\centering
\footnotesize
\setlength\tabcolsep{3pt} 
\begin{tabular}{lccccccccc}
\toprule[1pt]\midrule[0.3pt]
Dataset                   & TS2Vec                                & T-Loss                                & TNC                                   & TS-TCC                                & TST                               & DTW                                   & TimesNet* & \textbf{CoInception}                           & CoInception's Rank \\ \midrule
ArticularyWordRecognition & \textcolor{red}{\textbf{0.987}} & 0.943                                 & 0.973                                 & 0.953                                 & 0.977                             & \textcolor{red}{\textbf{0.987}} & 0.9       & \textcolor{red}{\textbf{0.987}} & 1    \\
AtrialFibrillation        & 0.2                                   & 0.133                                 & 0.133                                 & 0.267                                 & 0.067                             & 0.2                                   & 0.133     & \textcolor{red}{\textbf{0.333}} & 1    \\
BasicMotions              & 0.975                                 & \textcolor{red}{\textbf{1}}     & 0.975                                 & \textcolor{red}{\textbf{1}}     & 0.975                             & 0.975                                 & 0.9       & \textcolor{red}{\textbf{1}}     & 1    \\
CharacterTrajectories     & \textcolor{red}{\textbf{0.995}} & 0.993                                 & 0.967                                 & 0.985                                 & 0.975                             & 0.989                                 & 0.949     & 0.992                                 & 3    \\
Cricket                   & 0.972                                 & 0.972                                 & 0.958                                 & 0.917                                 & \textcolor{red}{\textbf{1}} & \textcolor{red}{\textbf{1}}     & 0.972     & 0.986                                 & 3    \\
\end{tabular}
\end{table*}

\subsection{Time Series Abnormally Detection}
\textbf{Additional details.}
In the preprocessing stage, we utilize the Augmented Dickey-Fuller (ADF) test, as done in \cite{tnc, ts2vec}, to determine the number of unit roots, denoted as $d$. Subsequently, the data is differenced $d$ times to mitigate any drifting effect, following the approach described in \cite{ts2vec}.

For the evaluation process, we adopt a similar protocol as presented in \cite{ts2vec, ren2019time, donut}, aimed at relaxing the point-wise detection constraint. Within this protocol, a small delay is allowed after the appearance of each anomaly point. Specifically, for minutely data, a maximum delay of $7$ steps is accepted, while for hourly data, a delay of $3$ steps is employed. If the detector correctly identifies the point within this delay, all points within the corresponding segment are considered correct; otherwise, they are deemed incorrect.

\subsection{Noise Resiliency Techniques Analysis}
\label{sec:app_jitter}
In our current sampling strategy, DWT low-pass filter acts as a denoising method treating input series $\mathbf{x}$ as a signal. While an alternative of introducing noise (like jittering) to ensure \emph{noise resiliency} is feasible, our preference for the DWT denoising technique stem from a realization. Jittering makes certain assumptions about the characteristics of the introduced noise, which may not be universally applicable to all time series or signals. In contrast, DWT denoising does not rely on such assumptions. The multiresolution breakdown in frequency achieved by DWT filters allows us to target specific high-frequency components prone to noise within the original signal. Through empirical analysis, we demonstrate the robustness of the DWT denoising technique compared to jittering.

We adhere to the commonly used parameters for the jittering augmentation technique described in \cite{jitter1, jitter2, jitter3}, where random noise is added from a Gaussian distribution with a mean ($\mu$) of 0 and a standard deviation ($\sigma$) of 0.03. To ensure a fair evaluation, we adopt all the settings as CoInception framework, making alterations only to the strategy employed in generating the perturbed series $\mathbf{x}$ during the training process. 
\renewcommand{\arraystretch}{1.0}
\begin{table*}[ht]
\setlength\doublerulesep{0.8pt}
\centering
\caption{Noise Resillency Techniques Comparison}
\label{tab:jitter}
\centering
\small
\setlength\tabcolsep{3pt} 
\begin{tabular}{llcc}
\toprule[1pt]\midrule[0.3pt]
\multicolumn{2}{c}{Task}                   & CoInception w. jittering & CoInception w. DWT filtering          \\ \midrule
                                    & Acc. &  0.656 (- 6.95\%)                        & \textcolor{red}{\textbf{0.705}} \\
\multirow{-2}{*}{Classification}    & AUC. &    0.727 (- 6.07\%)                      & \textcolor{red}{\textbf{0.774}} \\ \midrule
                                    & MSE  & 0.12 (- 49.12\%)        & \textcolor{red}{\textbf{0.061}} \\
\multirow{-2}{*}{Forecasting}       & MAE  & 0.262 (- 33.91\%)       & \textcolor{red}{\textbf{0.173}} \\ \midrule
                                    & F1   & 0.613 (- 20.28\%)       & \textcolor{red}{\textbf{0.769}} \\
                                    & P.   & 0.548 (- 30.63\%)        & \textcolor{red}{\textbf{0.790}} \\
\multirow{-3}{*}{Anomaly Detection} & R.   & 0.695 (- 7.08\%)        & \textcolor{red}{\textbf{0.748}} \\ \midrule[0.3pt]\bottomrule[1pt]
\end{tabular}
\end{table*}

Our experiments encompass all three main tasks, and the full results are reported in Table \ref{tab:jitter}. Across these three tasks, the DWT-based denoising technique consistently demonstrates its notable superiority over the jittering technique. It's worth reiterating that jittering assumes specific characteristics of the introduced noise, which may not universally apply to all time series or signals. In contrast, DWT denoising relies on an assumption generally applicable to natural signals: noisy elements typically manifest as high-frequency components within the original signal. We leave theoretical analysis and further exploration as open questions for our future research.

\subsection{Receptive Field Analysis}
\label{sec:app_receptive}
\begin{figure}[ht]
    \centering
    \includegraphics[width=0.7\columnwidth]{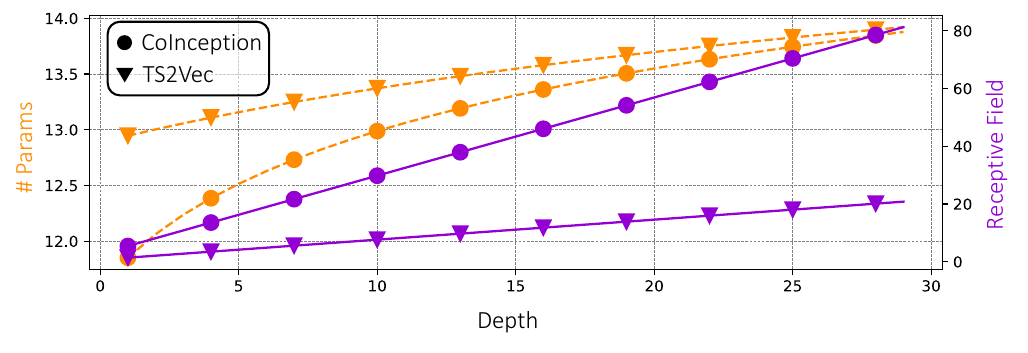}
    \caption{\textbf{Receptive Field Analysis}. The relation between models' depth with their number of parameters and their maximum receptive field. } \label{fig:scale_analysis}
\end{figure}

This experiment aims to investigate the scalability of the CoInception framework in comparison to the stacked Dilated Convolution network proposed in \cite{ts2vec}. We present a visualization of the relationship between the network depth, the number of parameters, and the maximum receptive fields of output timestamps in Figure \ref{fig:scale_analysis}.

The receptive field represents the number of input timestamps involved in calculating an output timestamp. The reported statistics for both the number of parameters and the receptive field are presented in logarithmic scale to ensure smoothness and a smaller number range.

As depicted in the figure, CoInception consistently exhibits a lower number of parameters compared to TS2Vec, across a network depth ranging from 1 to 30 layers. It is worth noting that the inclusion of a 30-layer CoInception framework in the visualization is purely for illustrative purposes, as we believe a much smaller depth is sufficient for the majority of time series datasets. In fact, we only utilize 3 layers for all datasets in the remaining sections. Furthermore, CoInception, with its multiple Basic units of varying filter lengths, can easily achieve very large receptive fields even with just a few layers.

\subsection{Clusterability Analysis}
\label{sec:clustering}
\begin{figure*}[th]
    \centering
    \begin{subfigure}{.32\linewidth}
    \centering
    \includegraphics[width=\textwidth]{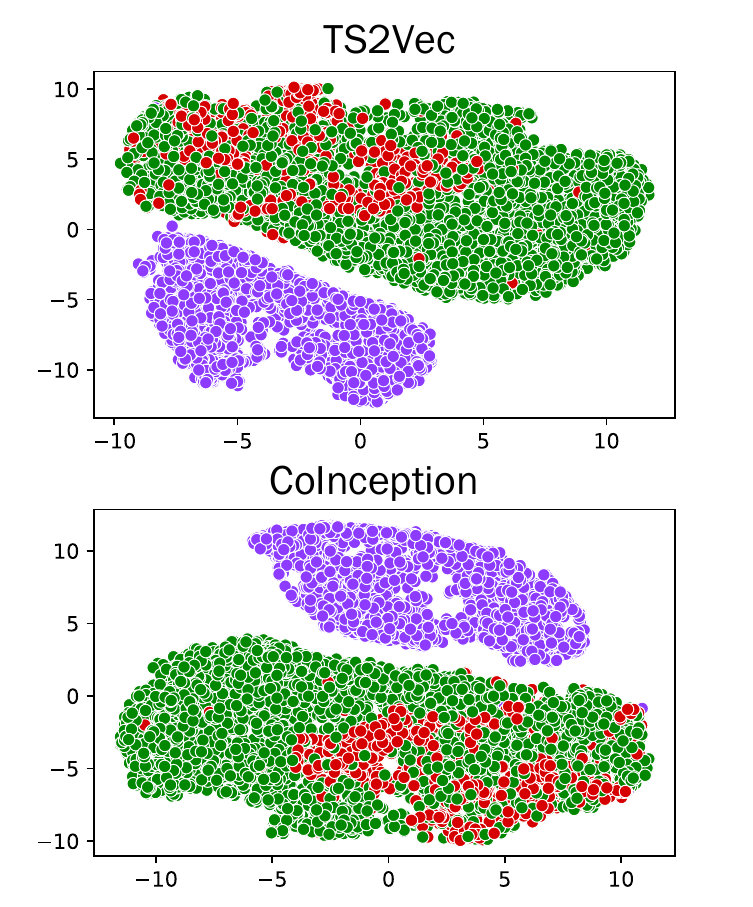}
    \caption{StarLightCurves}
    \label{fig:cluster_starlight}
    \end{subfigure}%
    \hfill
    \begin{subfigure}{.32\linewidth}
    \centering
    \includegraphics[width=\textwidth]{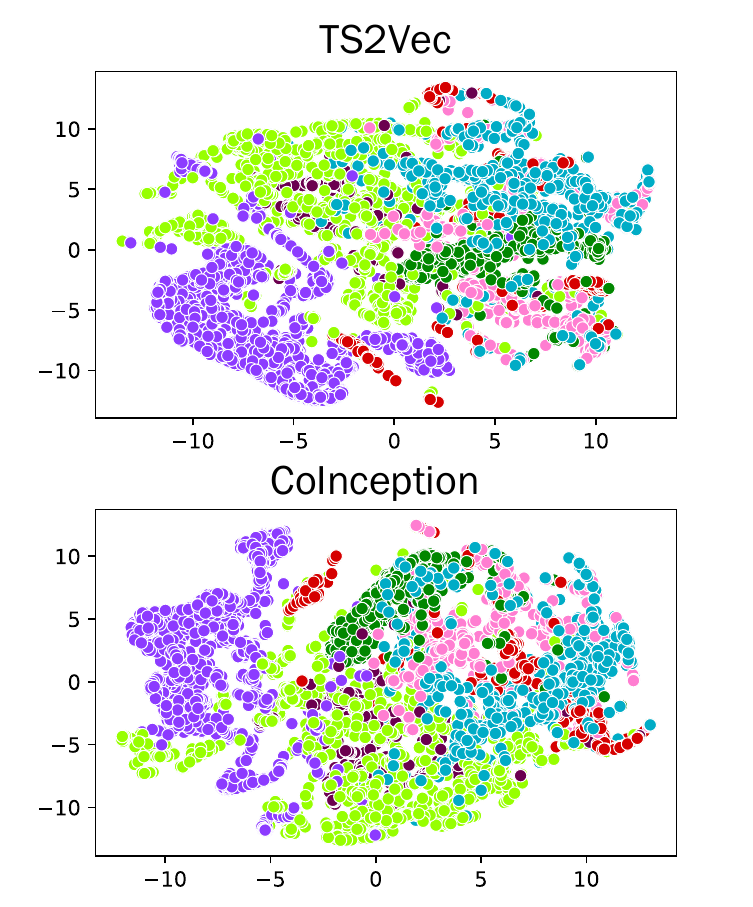}
    \caption{ElectricDevices}
    \label{fig:cluster_elec}
    \end{subfigure}
    \hfill
    \begin{subfigure}{.32\linewidth}
    \centering
    \includegraphics[width=\textwidth]{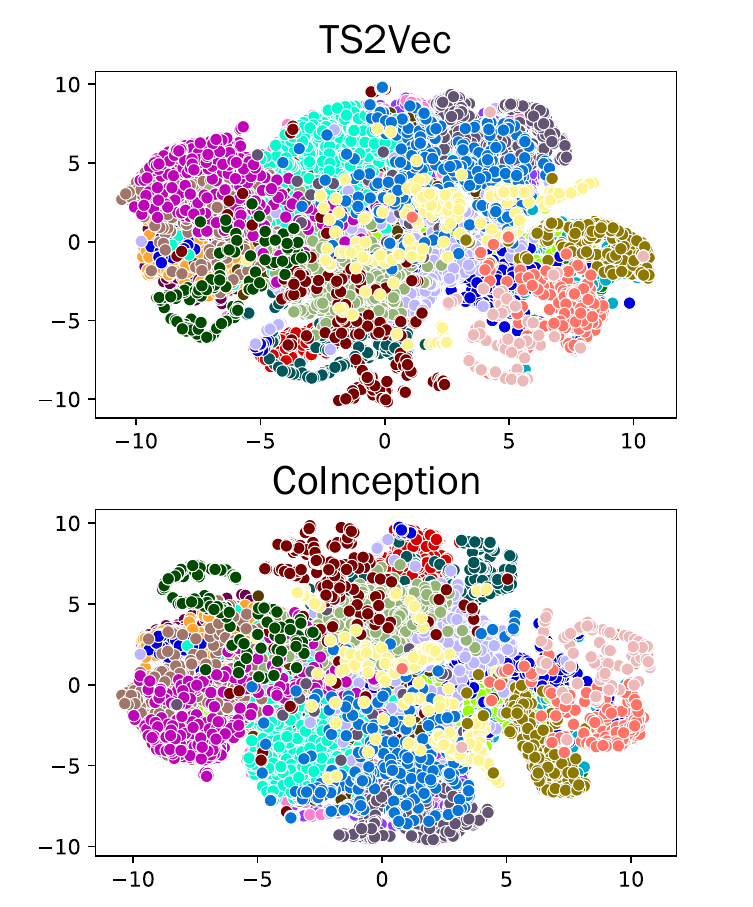}
    \caption{Crop}
    \label{fig:cluster_crop}
    \end{subfigure}
    \caption{Comparing clusterability of our CoInception with TS2Vec over three benchmark datasets in UCR 125 Repository.}  \label{fig:cluster}
\end{figure*}

Through this experiment, we test the clusterability of the learnt representations in the latent space. We visualize the feature representations with t-SNE proposed by Maaten and partners - \cite{t_sne} in two dimensional space. In the best scenario, the representations should be presented in latent space by groups of clusters, basing on their labels - their underlying states.

Figure \ref{fig:cluster} compares the distribution of representations learned by CoInception and TS2Vec in three dataset with greatest test set in UCR 128 repository. It is evident that the proposed CoInception does outperform the second best TS2Vec in terms of representation learning from the same hidden state. The clusters learnt by CoInception are more compact than those produced by TS2Vec, especially when the number of classes increase for \textit{ElectricDevices} or \textit{Crop} datasets. 

\subsection{Transferability Analysis}
\label{sec:app_transfer}
We assess the transferability of CoInception framework under all three tasks: forecasting, classification and anomaly detection. 

For the forecasting task, we evaluate the transferability of the CoInception framework using the following approach.
The ETT datasets \cite{informer} consist of power transformer data collected from July 2016 to July 2018. We focus on the small datasets, which include data from 2 stations, specifically load and oil temperature. ETTh1 and ETTh2 are datasets with a temporal granularity of 1 hour, corresponding to the two stations. Since these two datasets exhibit high correlation, we leverage transfer learning between them.
Initially, we perform the unsupervised learning step on the ETTh1 dataset, similar to the process used for forecasting assessment. Subsequently, the weights of the CoInception Encoder are frozen, and we utilize this pre-trained Encoder for training the forecasting framework, employing a Ridge Regression model, on the ETTh2 dataset. 
\renewcommand{\arraystretch}{1.0}
\begin{table*}[ht]
\setlength\doublerulesep{0.8pt}
\centering
\caption{Transferability analysis with time series forecasting task. 
}
\label{tab:transfer_forecasting}
\centering
\setlength\tabcolsep{3pt} 
\begin{tabular}{lccccc}
\toprule[1pt]\midrule[0.3pt]
                        & \multicolumn{5}{c}{Forecasting (ETTh1 -\textgreater ETTh2)}                                                                                                                                           \\ \cmidrule(lr){2-6} 
\multirow{-2}{*}{Model} & \multicolumn{1}{l}{24 Step}           & \multicolumn{1}{l}{48 Step}           & \multicolumn{1}{l}{168 Step}          & \multicolumn{1}{l}{336 Step}          & \multicolumn{1}{l}{720 Step}          \\ \midrule
TS2Vec                  & 0.090                                 & 0.124                                 & 0.208                                 & 0.213                                 & 0.214                                 \\
TS2Vec*                 & 0.100                                 & 0.143                                 & 0.236                                 & 0.223                                 & 0.217                                 \\
CoInception             & 0.086                                 & 0.119                                 & \textcolor{red}{ \textbf{0.185}} & \textcolor{red}{ \textbf{0.196}} & \textcolor{red}{ \textbf{0.209}} \\
CoInception*            & \textcolor{red}{ \textbf{0.084}} & \textcolor{red}{ \textbf{0.118}} & 0.188                                 & 0.201                                 & 0.211                                 \\ \midrule[0.3pt]\bottomrule[1pt]
\end{tabular}
\end{table*}
The detailed results are presented in Table \ref{tab:transfer_forecasting}. Overall, CoInception demonstrates its strong adaptability to the ETTh2 dataset, surpassing TS2Vec and even performing comparably to its own results in the regular forecasting setting.

For classification task, we follow the settings in \cite{t_loss}. We first train our Encoder unsupervisedly with training data from FordA dataset. Following, for each dataset in UCR repository, the SVM classifier is trained on top of the representations produced by the frozen CoInception Encoder with this dataset. 
Table \ref{tab:transfer_classification} provides a summary of the transferability results on the first 85 UCR datasets. Although CoInception exhibits lower performance compared to its own results in the regular classification setting in most datasets, its overall performance, as measured by the average accuracy, is still comparable to TS2Vec in its normal settings.

For anomaly detection, the settings are inherited from \cite{ren2019time, ts2vec}, and we have already presented the results with cold-start settings in the main manuscript.
\renewcommand{\arraystretch}{1.0}
\begin{table*}[ht]
\setlength\doublerulesep{0.8pt}
\centering
\caption{Transferability analysis for time series classification. 
}
\label{tab:transfer_classification}
\centering
\footnotesize
\setlength\tabcolsep{3pt} 
\begin{tabular}{lcccccc}
\toprule[1pt]\midrule[0.3pt]
Dataset                        & \multicolumn{1}{l}{TS2Vec}            & \multicolumn{1}{l}{TS2Vec$^*$}           & \multicolumn{1}{l}{T-Loss}            & \multicolumn{1}{l}{T-Loss$^*$}           & \multicolumn{1}{l}{CoInception}      & \multicolumn{1}{l}{CoInception$^*$}       \\ \midrule
Adiac                          & 0.762                                 & 0.783                                 & 0.760                                 & 0.716                                 & 0.767                                 & \textcolor{red}{\textbf{0.803}} \\
ArrowHead                      & 0.857                                 & 0.829                                 & 0.817                                 & 0.829                                 & \textcolor{red}{\textbf{0.863}} & 0.806                                 \\
Beef                           & \textcolor{red}{\textbf{0.767}} & 0.700                                 & 0.667                                 & 0.700                                 & 0.733                                 & 0.733                                 \\
BeetleFly                      & \textcolor{red}{\textbf{0.900}} & \textcolor{red}{\textbf{0.900}} & 0.800                                 & \textcolor{red}{\textbf{0.900}} & 0.850                                 & \textcolor{red}{\textbf{0.900}} \\
BirdChicken                    & 0.800                                 & 0.800                                 & \textcolor{red}{\textbf{0.900}} & 0.800                                 & \textcolor{red}{\textbf{0.900}} & 0.800                                 \\
Car                            & 0.833                                 & 0.817                                 & 0.850                                 & 0.817                                 & 0.867                                 & \textcolor{red}{\textbf{0.883}} \\
CBF                            & \textcolor{red}{\textbf{1.000}} & \textcolor{red}{\textbf{1.000}} & 0.988                                 & 0.994                                 & \textcolor{red}{\textbf{1.000}} & 0.997                                 \\
ChlorineConcentration          & \textcolor{red}{\textbf{0.832}} & 0.802                                 & 0.688                                 & 0.782                                 & 0.813                                 & 0.814                                 \\
CinCECGTorso                   & \textcolor{red}{\textbf{0.827}} & 0.738                                 & 0.638                                 & 0.740                                 & 0.765                                 & 0.772                                 \\
Coffee                         & \textcolor{red}{\textbf{1.000}} & \textcolor{red}{\textbf{1.000}} & \textcolor{red}{\textbf{1.000}} & \textcolor{red}{\textbf{1.000}} & \textcolor{red}{\textbf{1.000}} & \textcolor{red}{\textbf{1.000}} \\
Computers                      & 0.660                                 & 0.660                                 & 0.648                                 & 0.628                                 & \textcolor{red}{\textbf{0.688}} & 0.668                                 \\
CricketX                       & 0.782                                 & 0.767                                 & 0.682                                 & 0.777                                 & \textcolor{red}{\textbf{0.805}} & 0.767                                 \\
CricketY                       & 0.749                                 & 0.746                                 & 0.667                                 & 0.767                                 & \textcolor{red}{\textbf{0.818}} & 0.751                                 \\
CricketZ                       & 0.792                                 & 0.772                                 & 0.656                                 & 0.764                                 & \textcolor{red}{\textbf{0.808}} & 0.762                                 \\
DiatomSizeReduction            & 0.984                                 & 0.961                                 & 0.974                                 & \textcolor{red}{\textbf{0.993}} & 0.984                                 & 0.977                                 \\
DistalPhalanxOutlineCorrect    & 0.761                                 & 0.757                                 & 0.764                                 & 0.768                                 & \textcolor{red}{\textbf{0.779}} & 0.775                                 \\
DistalPhalanxOutlineAgeGroup   & 0.727                                 & \textcolor{red}{\textbf{0.748}} & 0.727                                 & 0.734                                 & \textcolor{red}{\textbf{0.748}} & 0.741                                 \\
DistalPhalanxTW                & 0.698                                 & 0.669                                 & 0.669                                 & 0.676                                 & \textcolor{red}{\textbf{0.705}} & 0.698                                 \\
Earthquakes                    & \textcolor{red}{\textbf{0.748}} & \textcolor{red}{\textbf{0.748}} & \textcolor{red}{\textbf{0.748}} & \textcolor{red}{\textbf{0.748}} & \textcolor{red}{\textbf{0.748}} & \textcolor{red}{\textbf{0.748}} \\
ECG200                         & \textcolor{red}{\textbf{0.920}} & 0.910                                 & 0.830                                 & 0.900                                 & \textcolor{red}{\textbf{0.920}} & \textcolor{red}{\textbf{0.920}} \\
ECG5000                        & 0.935                                 & 0.935                                 & 0.940                                 & 0.936                                 & \textcolor{red}{\textbf{0.944}} & 0.942                                 \\
ECGFiveDays                    & \textcolor{red}{\textbf{1.000}} & \textcolor{red}{\textbf{1.000}} & \textcolor{red}{\textbf{1.000}} & \textcolor{red}{\textbf{1.000}} & \textcolor{red}{\textbf{1.000}} & \textcolor{red}{\textbf{1.000}} \\
ElectricDevices                & 0.721                                 & 0.714                                 & 0.676                                 & 0.732                                 & \textcolor{red}{\textbf{0.741}} & 0.722                                 \\
FaceAll                        & 0.771                                 & 0.786                                 & 0.734                                 & 0.802                                 & \textcolor{red}{\textbf{0.842}} & 0.821                                 \\
FaceFour                       & 0.932                                 & 0.898                                 & 0.830                                 & 0.875                                 & \textcolor{red}{\textbf{0.955}} & 0.807                                 \\
FacesUCR                       & 0.924                                 & \textcolor{red}{\textbf{0.928}} & 0.835                                 & 0.918                                 & \textcolor{red}{\textbf{0.928}} & 0.923                                 \\
FiftyWords                     & 0.771                                 & 0.785                                 & 0.745                                 & 0.780                                 & 0.778                                 & \textcolor{red}{\textbf{0.804}} \\
Fish                           & 0.926                                 & 0.949                                 & \textcolor{red}{\textbf{0.960}} & 0.880                                 & 0.954                                 & 0.943                                 \\
FordA                          & 0.936                                 & 0.936                                 & 0.927                                 & 0.935                                 & 0.930                                 & \textcolor{red}{\textbf{0.943}} \\
FordB                          & 0.794                                 & 0.779                                 & 0.798                                 & \textcolor{red}{\textbf{0.810}} & 0.802                                 & 0.796                                 \\
GunPoint                       & 0.980                                 & \textcolor{red}{\textbf{0.993}} & 0.987                                 & \textcolor{red}{\textbf{0.993}} & 0.987                                 & 0.987                                 \\
Ham                            & 0.714                                 & 0.714                                 & 0.533                                 & 0.695                                 & \textcolor{red}{\textbf{0.810}} & 0.648                                 \\
HandOutlines                   & 0.922                                 & 0.919                                 & 0.919                                 & 0.922                                 & \textcolor{red}{\textbf{0.935}} & 0.930                                 \\
Haptics                        & \textcolor{red}{\textbf{0.526}} & \textcolor{red}{\textbf{0.526}} & 0.474                                 & 0.455                                 & 0.510                                 & 0.513                                 \\
Herring                        & \textcolor{red}{\textbf{0.641}} & 0.594                                 & 0.578                                 & 0.578                                 & 0.594                                 & 0.609                                 \\
InlineSkate                    & 0.415                                 & \textcolor{red}{\textbf{0.465}} & 0.444                                 & 0.447                                 & 0.424                                 & 0.453                                 \\
InsectWingbeatSound            & 0.630                                 & 0.603                                 & 0.599                                 & 0.623                                 & \textcolor{red}{\textbf{0.634}} & 0.630                                 \\
ItalyPowerDemand               & 0.925                                 & 0.957                                 & 0.929                                 & 0.925                                 & 0.962                                 & \textcolor{red}{\textbf{0.963}} \\
LargeKitchenAppliances         & 0.845                                 & 0.861                                 & 0.765                                 & 0.848                                 & \textcolor{red}{\textbf{0.893}} & 0.787                                 \\
Lightning2                     & 0.869                                 & \textcolor{red}{\textbf{0.918}} & 0.787                                 & \textcolor{red}{\textbf{0.918}} & 0.902                                 & 0.852                                 \\
Lightning7                     & \textcolor{red}{\textbf{0.863}} & 0.781                                 & 0.740                                 & 0.795                                 & 0.836                                 & 0.808                                 \\
Mallat                         & 0.914                                 & 0.956                                 & 0.916                                 & 0.964                                 & 0.953                                 & \textcolor{red}{\textbf{0.966}} \\
Meat                           & 0.950                                 & \textcolor{red}{\textbf{0.967}} & 0.867                                 & 0.950                                 & \textcolor{red}{\textbf{0.967}} & \textcolor{red}{\textbf{0.967}} \\
MedicalImages                  & 0.789                                 & 0.784                                 & 0.725                                 & 0.784                                 & \textcolor{red}{\textbf{0.795}} & 0.792                                 \\
MiddlePhalanxOutlineCorrect    & \textcolor{red}{\textbf{0.838}} & 0.794                                 & 0.787                                 & 0.814                                 & 0.832                                 & \textcolor{red}{\textbf{0.838}} \\
MiddlePhalanxOutlineAgeGroup   & 0.636                                 & 0.649                                 & 0.623                                 & 0.656                                 & 0.656                                 & \textcolor{red}{\textbf{0.662}} \\
MiddlePhalanxTW                & 0.584                                 & 0.597                                 & 0.584                                 & \textcolor{red}{\textbf{0.610}} & 0.604                                 & \textcolor{red}{\textbf{0.610}} \\
MoteStrain                     & 0.861                                 & 0.847                                 & 0.823                                 & 0.871                                 & \textcolor{red}{\textbf{0.873}} & 0.822                                 \\
NonInvasiveFetalECGThorax1     & 0.930                                 & 0.946                                 & 0.925                                 & 0.910                                 & 0.919                                 & \textcolor{red}{\textbf{0.947}} \\
NonInvasiveFetalECGThorax2     & 0.938                                 & \textcolor{red}{\textbf{0.955}} & 0.930                                 & 0.927                                 & 0.942                                 & 0.950                                 \\
OliveOil                       & \textcolor{red}{\textbf{0.900}} & \textcolor{red}{\textbf{0.900}} & \textcolor{red}{\textbf{0.900}} & \textcolor{red}{\textbf{0.900}} & \textcolor{red}{\textbf{0.900}} & \textcolor{red}{\textbf{0.900}} \\
OSULeaf                        & 0.851                                 & \textcolor{red}{\textbf{0.868}} & 0.736                                 & 0.831                                 & 0.835                                 & 0.777                                 \\
PhalangesOutlinesCorrect       & 0.809                                 & 0.794                                 & 0.784                                 & 0.801                                 & \textcolor{red}{\textbf{0.818}} & 0.800                                 \\
Phoneme                        & \textcolor{red}{\textbf{0.312}} & 0.260                                 & 0.196                                 & 0.289                                 & 0.310                                 & 0.294                                 \\
\end{tabular}
\end{table*}
\renewcommand{\arraystretch}{1.0}
\begin{table*}[ht]
\setlength\doublerulesep{0.8pt}
\centering
\footnotesize
\setlength\tabcolsep{3pt} 
\begin{tabular}{lrrrrrr}
\toprule[1pt]\midrule[0.3pt]
Dataset                        & \multicolumn{1}{l}{TS2Vec}            & \multicolumn{1}{l}{TS2Vec$^*$}           & \multicolumn{1}{l}{T-Loss}            & \multicolumn{1}{l}{T-Loss$^*$}           & \multicolumn{1}{l}{CoInception}      & \multicolumn{1}{l}{CoInception$^*$}       \\ \midrule
Plane                          & \textcolor{red}{\textbf{1.000}} & 0.981                                 & 0.981                                 & 0.990                                 & \textcolor{red}{\textbf{1.000}} & \textcolor{red}{\textbf{1.000}} \\
ProximalPhalanxOutlineCorrect  & 0.887                                 & 0.876                                 & 0.869                                 & 0.859                                 & \textcolor{red}{\textbf{0.911}} & 0.893                                 \\
ProximalPhalanxOutlineAgeGroup & 0.834                                 & 0.844                                 & 0.839                                 & \textcolor{red}{\textbf{0.854}} & 0.849                                 & 0.844                                 \\
ProximalPhalanxTW              & \textcolor{red}{\textbf{0.824}} & 0.805                                 & 0.785                                 & \textcolor{red}{\textbf{0.824}} & \textcolor{red}{\textbf{0.824}} & 0.820                                 \\
RefrigerationDevices           & 0.589                                 & 0.557                                 & 0.555                                 & 0.517                                 & 0.597                                 & \textcolor{red}{\textbf{0.635}} \\
ScreenType                     & 0.411                                 & 0.421                                 & 0.384                                 & 0.413                                 & 0.413                                 & \textcolor{red}{\textbf{0.469}} \\
ShapeletSim                    & \textcolor{red}{\textbf{1.000}} & \textcolor{red}{\textbf{1.000}} & 0.517                                 & 0.817                                 & 0.994                                 & \textcolor{red}{\textbf{1.000}} \\
ShapesAll                      & \textcolor{red}{\textbf{0.902}} & 0.877                                 & 0.837                                 & 0.875                                 & 0.898                                 & 0.863                                 \\
SmallKitchenAppliances         & 0.731                                 & 0.747                                 & 0.731                                 & 0.715                                 & \textcolor{red}{\textbf{0.792}} & 0.717                                 \\
SonyAIBORobotSurface1          & 0.903                                 & 0.884                                 & 0.840                                 & 0.897                                 & \textcolor{red}{\textbf{0.908}} & 0.903                                 \\
SonyAIBORobotSurface2          & 0.871                                 & 0.872                                 & 0.832                                 & 0.934                                 & 0.939                                 & \textcolor{red}{\textbf{0.940}} \\
StarLightCurves                & 0.969                                 & 0.967                                 & 0.968                                 & 0.965                                 & 0.971                                 & \textcolor{red}{\textbf{0.974}} \\
Strawberry                     & 0.962                                 & 0.962                                 & 0.946                                 & 0.946                                 & \textcolor{red}{\textbf{0.970}} & \textcolor{red}{\textbf{0.970}} \\
SwedishLeaf                    & 0.941                                 & 0.931                                 & 0.925                                 & 0.931                                 & 0.950                                 & \textcolor{red}{\textbf{0.957}} \\
Symbols                        & \textcolor{red}{\textbf{0.976}} & 0.973                                 & 0.945                                 & 0.965                                 & 0.970                                 & 0.961                                 \\
SyntheticControl               & \textcolor{red}{\textbf{0.997}} & \textcolor{red}{\textbf{0.997}} & 0.977                                 & 0.983                                 & \textcolor{red}{\textbf{0.997}} & 0.990                                 \\
ToeSegmentation1               & 0.917                                 & 0.947                                 & 0.899                                 & \textcolor{red}{\textbf{0.952}} & 0.943                                 & 0.947                                 \\
ToeSegmentation2               & 0.892                                 & \textcolor{red}{\textbf{0.946}} & 0.900                                 & 0.885                                 & 0.908                                 & 0.900                                 \\
Trace                          & \textcolor{red}{\textbf{1.000}} & \textcolor{red}{\textbf{1.000}} & \textcolor{red}{\textbf{1.000}} & \textcolor{red}{\textbf{1.000}} & \textcolor{red}{\textbf{1.000}} & \textcolor{red}{\textbf{1.000}} \\
TwoLeadECG                     & 0.986                                 & \textcolor{red}{\textbf{0.999}} & 0.993                                 & 0.997                                 & 0.998                                 & \textcolor{red}{\textbf{0.999}} \\
TwoPatterns                    & \textcolor{red}{\textbf{1.000}} & 0.999                                 & 0.992                                 & \textcolor{red}{\textbf{1.000}} & \textcolor{red}{\textbf{1.000}} & \textcolor{red}{\textbf{1.000}} \\
UWaveGestureLibraryX           & 0.795                                 & 0.818                                 & 0.784                                 & 0.811                                 & 0.817                                 & \textcolor{red}{\textbf{0.820}} \\
UWaveGestureLibraryY           & 0.719                                 & \textcolor{red}{\textbf{0.739}} & 0.697                                 & 0.735                                 & \textcolor{red}{\textbf{0.739}} & 0.738                                 \\
UWaveGestureLibraryZ           & 0.770                                 & 0.757                                 & 0.729                                 & 0.759                                 & \textcolor{red}{\textbf{0.771}} & 0.754                                 \\
UWaveGestureLibraryAll         & 0.930                                 & 0.918                                 & 0.865                                 & 0.941                                 & 0.937                                 & \textcolor{red}{\textbf{0.956}} \\
Wafer                          & 0.998                                 & 0.997                                 & 0.995                                 & 0.993                                 & \textcolor{red}{\textbf{0.999}} & 0.998                                 \\
Wine                           & 0.870                                 & 0.759                                 & 0.685                                 & 0.870                                 & \textcolor{red}{\textbf{0.907}} & \textcolor{red}{\textbf{0.907}} \\
WordSynonyms                   & 0.676                                 & 0.693                                 & 0.641                                 & \textcolor{red}{\textbf{0.704}} & 0.683                                 & 0.691                                 \\
Worms                          & 0.701                                 & \textcolor{red}{\textbf{0.753}} & 0.688                                 & 0.714                                 & 0.740                                 & 0.701                                 \\
WormsTwoClass                  & 0.805                                 & 0.688                                 & 0.753                                 & \textcolor{red}{\textbf{0.818}} & \textcolor{red}{\textbf{0.818}} & 0.779                                 \\
Yoga                           & \textcolor{red}{\textbf{0.887}} & 0.855                                 & 0.828                                 & 0.878                                 & 0.882                                 & 0.854                                 \\ \midrule
Avg. (first 85 datasets)       & 0.829                                 & 0.824                                 & 0.786                                 & 0.821                                 & \textcolor{red}{\textbf{0.841}} & \textcolor{red}{\textbf{0.829}} \\
\midrule[0.3pt]\bottomrule[1pt]
\end{tabular}
\end{table*}
\subsection{Additional Ablation Analysis}
\label{sec:app_ablation}
Through this experiment, we further analyze the effect of each individual contribution within Inception block. To be specific, three variances are adopts: \textbf{(2a)} We replace Aggregator with simple concatenation operation, and add Bottleneck layer followed the design in \cite{timesnet}; \textbf{(2b)} Dilated Convolution are turned into normal 1D Convolution layer; \textbf{(2c)} Skip connections between Basic Units of different layers are removed. The results are provided in Table \ref{tab:ablation_2}. Overall, while different ablations show the greater detrimental levels in different tasks, we consistently notice a decline in the whole performance when any suggested alteration is excluded or substituted. This pattern indicates the positive impact of each change on the robustness of the CoInception framework.

\begin{figure}
\setlength\doublerulesep{0.7pt}
\centering
\caption{Ablation analysis for Inception block of CoInception framework.}
\label{tab:ablation_2}
\centering
\small
\setlength\tabcolsep{3pt} 
\begin{tabular}{lcccc}
\toprule[1pt]\midrule[0.3pt]
     & CoInception (2a)                         & CoInception (2b)  & CoInception (2c)   & \textcolor{red}{\textbf{CoInception}} \\ \midrule
\multicolumn{5}{l}{\textit{Classification:}}                                                                                        \\
Acc. & \textcolor{blue}{0.671 \textbf{(- 4.82\%)}} & 0.571 \textbf{(- 19.00\%)} & 0.654 \textbf{(- 7.23\%}) & \textcolor{red}{0.705}                                       \\
AUC. & \textcolor{blue}{0.734 \textbf{(- 5.16\%)}}                        & 0.635 \textbf{(- 17.95\%)} & 0.719 \textbf{(- 7.11\%)} & \textcolor{red}{0.774}                                       \\ \midrule
\multicolumn{5}{l}{\textit{Forecasting:}}                                                                                           \\
MSE  & 0.068 \textbf{(- 10.29\%)}                        & 0.064 \textbf{(- 4.68\%)} & \textcolor{red}{0.060 \textbf{(+ 1.66\%)}}  & \textcolor{blue}{0.061}                                     \\
MAE  & 0.186 \textbf{(- 6.98\%)}                        & 0.178 \textbf{(- 2.80\%)} & \textcolor{red}{0.172 \textbf{(+ 0.58\%)}}  & \textcolor{blue}{0.173}                                       \\ \midrule
\multicolumn{5}{l}{\textit{Anomaly Detection:}}                                                                                     \\
F1   & 0.648 \textbf{(- 15.73\%)}                       & 0.647 \textbf{(- 15.86\%)} & \textcolor{blue}{0.653 \textbf{(-15.08\%)}}  & \textcolor{red}{0.769}                                       \\
P.   & 0.626 \textbf{(- 20.75\%)}                       & \textcolor{blue}{0.639 \textbf{(- 19.11\%)}} & 0.617 \textbf{(-21.89\%)}  & \textcolor{red}{0.790}                                       \\
R.   & 0.671 \textbf{(- 10.29\%)}                        & 0.656 \textbf{(- 12.29\%)} & \textcolor{blue}{0.694 \textbf{(- 7.21\%)}}  & \textcolor{red}{0.748}                                       \\ \midrule[0.3pt]\bottomrule[1pt]
\end{tabular}
\end{figure}

\section{Additional dicussions regarding CoInception}
This section is dedicated to discussing certain limitations and potential drawbacks of the CoInception framework. These insights aim to assist readers in determining suitable applications for CoInception.

About the sampling strategy based on DWT, we implicitly limit the noise being targeted in this study to be high-frequency. By removing the high-frequency components, the filter helps to smooth out the signal and eliminate rapid fluctuations caused by noise, while better revealing the underlying trends or slow-varying patterns in the time series. However, this strategy does not necessarily create an ideal noise-free signal of the series. In the circumstance where the dataset is either completely free of noise or inherently possesses noise predominantly in the low-frequency spectrum (such as a drifting effect \cite{lfreq_1, lfreq_2}), our proposed strategy might not offer significant benefits in managing those noisy signals.

About the encoder architecture, while the current design aligns with our main criteria of reaching both efficiency and effectiveness, it comes with a potential trade-off. With the use of Inception idea to automate the choice of scaling dilation factors, the problem of optimizing the number of layers used remains to be answered. This problem is also related to efficiency-effectiveness trade-off, hence it still needs extra effort to determine the number of layers used in the architecture. While we currently limit and fix our framework with 3 layers, we make no claim about the optimal number of layers to be used, but should be fine-tuned instead depending on the tasks or datasets specifically. We would consider this factor for a future study.

\end{document}